%% file: iclr2026_conference.tex
\documentclass{article} 
\usepackage{iclr2026_conference,times}
\usepackage{hyperref}       
\usepackage{url}            
\usepackage{booktabs}       
\usepackage{amsfonts}       
\usepackage{nicefrac}       
\usepackage{microtype}      
\usepackage{array}
\usepackage{multirow}
\usepackage{adjustbox}
\usepackage{amsmath} 
\usepackage{amsthm}
\usepackage[table]{xcolor} 
\usepackage{makecell}
\usepackage{amssymb}
\usepackage{mathtools}
\usepackage{arydshln}
\usepackage[accsupp]{axessibility}
\usepackage{multirow}
\usepackage{multicol}
\usepackage{pgfplots} 
\usepackage{overpic} 
\usepackage{enumitem}
\usepackage{pifont} 
\usepackage{algorithm}
\usepackage{algorithmic}
\usepackage{setspace}
\usepackage{bbding}
\usepackage{graphicx}    
\usepackage{subcaption}  
\usepackage{wrapfig}
\iclrfinalcopy
\input{math_commands.tex}

\newcommand{\blutiny}[1]{\tiny \textbf{\textcolor[HTML]{4C4CFF}{#1}}}
\newcommand{\op}[1]{\operatorname{#1}}

\newcommand{\cmark}{\ding{51}}%
\newcommand{\xmark}{\ding{55}}%
\usepackage[table]{xcolor}
\newcommand{\tabincell}[2]{\begin{tabular}{@{}#1@{}}#2\end{tabular}}
\definecolor{color3}{RGB}{255, 255, 204}
\definecolor{color2}{RGB}{255, 229, 204}
\definecolor{color1}{RGB}{229, 204, 255} 
\title{TP-Spikformer: Token Pruned Spiking\\ Transformer}

\author{Wenjie Wei$^1$,\; Xiaolong Zhou$^1$, Malu Zhang$^{1,2}$\thanks{Corresponding author: maluzhang@uestc.edu.cn},\; Ammar Belatreche$^3$,\; Qian Sun$^1$,\\ \textbf{Yimeng Shan$^{1}$,}\; \textbf{Dehao Zhang$^1$,}\; \textbf{Zijian Zhou$^1$,}\; \textbf{Zeyu Ma$^1$,}\; \textbf{Yang Yang$^1$,}\; \textbf{Haizhou Li$^{2,4}$} \\
  $^1$University of Electronic Science and Technology of China,~$^2$Shenzhen Loop Area Institute,\\
  $^3$Northumbria University,~$^4$The Chinese University of Hong Kong, Shenzhen (CUHK-Shenzhen)}

\begin{document}

\maketitle

\begin{abstract}
Spiking neural networks (SNNs) offer an energy-efficient alternative to traditional neural networks due to their event-driven computing paradigm. However, recent advancements in spiking transformers have focused on improving accuracy with large-scale architectures, which require significant computational resources and limit deployment on resource-constrained devices. In this paper, we propose a simple yet effective token pruning method for spiking transformers, termed TP-Spikformer, that reduces storage and computational overhead while maintaining competitive performance. Specifically, we first introduce a heuristic spatiotemporal information-retaining criterion that comprehensively evaluates tokens' importance, assigning higher scores to informative tokens for retention and lower scores to uninformative ones for pruning. Based on this criterion, we propose an information-retaining token pruning framework that employs a block-level early stopping strategy for uninformative tokens, instead of removing them outright. This also helps preserve more information during token pruning. We demonstrate the effectiveness, efficiency and scalability of TP-Spikformer through extensive experiments across diverse architectures, including Spikformer, QKFormer and Spike-driven Transformer V1 and V3, and a range of tasks such as image classification, object detection, semantic segmentation and event-based object tracking. Particularly, TP-Spikformer performs well in a training-free manner. These results reveal its potential as an efficient and practical solution for deploying SNNs in real-world applications with limited computational resources.
\end{abstract}

\section{Introduction}
Spiking Neural Networks (SNNs) have emerged as a promising energy-efficient solution for next-generation machine intelligence \cite{gerstner2002spiking,izhikevich2003simple}. In SNNs, the discrete binary spike serves as the fundamental information carrier and is conveyed event-drivenly.
This unique computing paradigm allows only a subset of neurons to be activated and engage in synaptic accumulation operations, achieving significant computational efficiency \cite{pfeiffer2018deep,roy2019towards,li2024brain}. In addition, the sparse event-driven nature of SNNs has spurred the development of neuromorphic hardware, such as TrueNorth \cite{akopyan2015truenorth} and Loihi \cite{davies2018loihi}, further harnessing their potential for energy efficiency. 
Despite the significant efficiency advantages, the limited performance of SNNs presents challenges to their widespread applications.

Building on the success of Transformer models across various fields \cite{devlin2019bert,dosovitskiy2020image}, researchers have integrated them with SNNs, such as Spikformer \cite{zhouspikformer}, QKFormer\cite{zhang2024qkformer}, Spike-driven transformer(SDT)-V1 \cite{yao2023spike}, V2 \cite{yaospike} and V3 \cite{yao2025scaling}, leading to significant performance improvements on large and complex benchmarks. However, these gains come at the expense of a large number of model parameters and high computational complexity.
For example, the recently introduced SDT-V3 \cite{yao2025scaling} achieves 86.2\% accuracy on ImageNet. Yet, this model contains 173 million parameters, requires 1384MB of memory, and performs 28.4 billion synaptic operations per second during inference. These present significant challenges for deploying Transformer-based SNNs on resource-limited scenarios \cite{qiu2025quantized,zhan2025sfedca}.

Researchers have made significant efforts to compress large-scale spiking transformers, including techniques of quantization \cite{cao2025binary,wei2024q}, network architecture search \cite{wang2024autost,che2024auto,zhang2025spike}, and pruning \cite{zhuge2024towards,zhou2024spatial}. 
Among these, token pruning dynamically reduces the number of tokens processed in each block during inference, enhancing both storage and computational efficiency. The underlying principle behind it is that, in visual tasks, the final prediction typically relies on only a subset of the tokens. This allows us to selectively remove certain tokens, accelerating inference while maintaining competitive accuracy \cite{rao2021dynamicvit,yin2022vit}.
Therefore, token pruning poses a promising solution for the efficient deployment of spiking transformers, particularly in edge scenarios.
However, current token pruning methods in SNNs suffer from two major limitations \cite{zhuge2024towards,liu2024sparsespikformer,kang2024snn}. First, most approaches modify the original structure when applied to spiking transformers, such as introducing tokens, adding trainable modules, or altering network connections. Second, these methods typically require retraining the model, resulting in large training costs. These issues raise the application costs and reduce their generalizability.

In this paper, we propose a simple yet effective token pruning approach for spiking transformer (TP-Spikformer), aiming to compress its storage and accelerate its computation while maintaining competitive performance. 
We first propose a {heuristic spatiotemporal information-retaining token pruning criterion} (IRToP), where informative tokens are assigned higher scores to retain and uninformative tokens are given lower scores to prune. 
Based on this criterion, we design an information-retaining token pruning architecture (IR-Arc), which achieves compression and acceleration by applying a block-level early stopping strategy for uninformative tokens, rather than direct dropping. This also helps retain more information during token pruning.
The token pruning results of TP-Spikformer are depicted in Figure \ref{fig:tokendrop}, and our main contributions are summarized as follows:
\begin{figure}[t]
  \centering
  \includegraphics[scale=0.56]{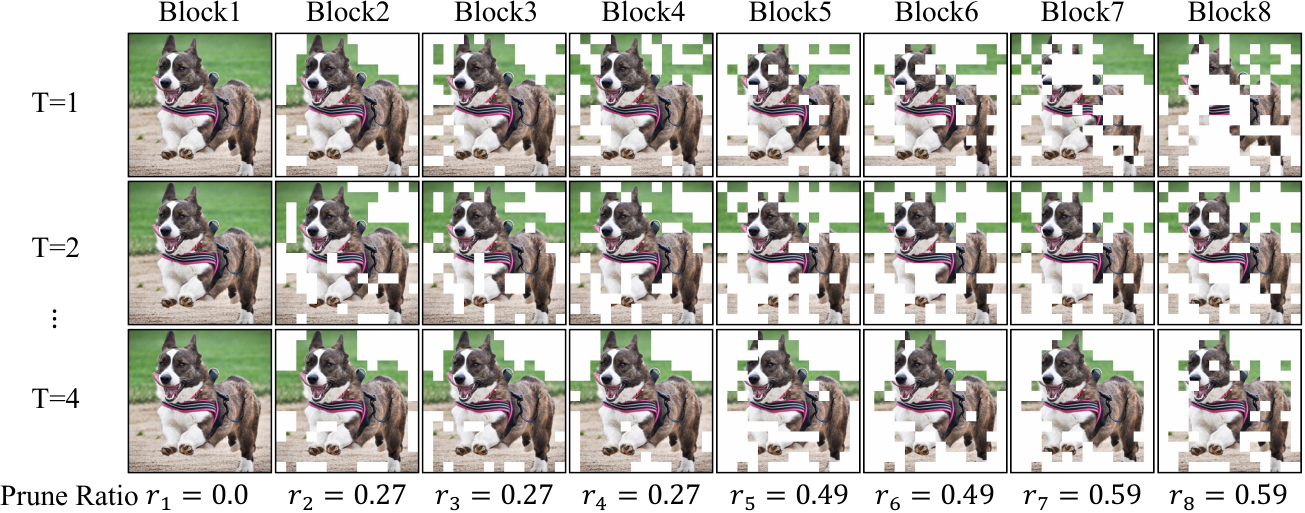}
  \caption{Visualization of token pruning across time step and block with our method. Experiments are conducted on SDT-V1-8-512, and white areas are pruned tokens. 
  }
  \label{fig:tokendrop}
  \vspace{-5pt}
\end{figure}
\begin{itemize}
    \item We propose a heuristic spatiotemporal information-retaining criterion for token pruning, termed IRToP. Spatially, IRToP recognizes tokens that differ significantly from neighbors as more distinctive. Temporally, IRToP identifies tokens with greater variation across adjacent time steps as carriers of richer temporal information. By integrating both aspects, IRToP effectively identifies informative tokens and assigns them higher retention priority.

    \item We propose an information-retention token pruning architecture, named IR-Arc, where informative tokens undergo complete forward computation but uninformative tokens implement a block-level early stopping strategy, reducing storage and computation overhead effectively. IR-Arc makes TP-Spikformer exhibit high versatility and requires no training from scratch, attaining competitive performance even under zero fine-tuning conditions.


    \item We select a variety of architectures for our experiments, including the feature-map invariant Spikformer and SDT-V1, the feature pyramid-based QKFormer, and the advanced SDT-V3. Additionally, we assess TP-Spikformer on multiple tasks such as classification, segmentation, detection, and tracking. Through validation across multiple architectures and tasks, we demonstrate the effectiveness, efficiency, and scalability of our method.
    
\end{itemize}

\section{Related work}

\textbf{Spiking transformer.}
Spikformer pioneers self-attention and direct Transformer training in SNNs \cite{zhouspikformer}, which eliminates float multiplication in attention via spike-based Q, K, and V.
Spikformer V2 explores masked image modeling in spiking transformers, achieving 81.1\% accuracy on ImageNet with just 1 time step \cite{zhou2024spikformer}.
SpikingResformer proposes a dual-spike self-attention and combines it with a ResNet-based architecture, improving performance with reduced parameters \cite{shi2024spikingresformer}.
QKFormer uses spike-based Q and K for attention computation and introduces spiking patch embeddings with deformable shortcuts, achieving milestone results on multiple datasets \cite{zhou2024qkformer}.
In the SNN community, the series of Spike-driven Transformer has gained notable attention.
SDT-V1 pioneers spike-driven computation in spiking transformers, converting spike-related matrix multiplications to efficient addition operations \cite{yao2023spike}.
SDT-V2 extends SDT-V1 into a meta architecture, exploring structure design, spike-driven attention, and skip connection to enhance performance \cite{yaospike}.
SDT-V3 optimizes spiking neuron firing patterns and designs an efficient Transformer \cite{yao2025scaling}.
Despite substantial accuracy improvements, SNNs' inherent energy efficiency is undermined, limiting their deployment in resource-limited scenarios.

\textbf{Token pruning in spiking transformer.}
SparseSpikformer proposes a hybrid pruning framework operating at both weight and token levels, removing unimportant background tokens based on neurons' spike firing rates \cite{liu2024sparsespikformer}. However, it has two limitations: it relies on firing rate for token importance without leveraging SNNs' temporal characteristics, and its validation is limited to a single architecture and small-scale datasets, leaving its scalability to other architectures and benchmarks unexplored.
AT-SNN adopts an adaptive computation time (ACT) mechanism to mask unimportant tokens using Halting Scores during training, followed by a similarity-based token merging strategy to reduce computational overhead \cite{kang2024snn}. However, ACT introduces additional parameters requiring retraining, and AT-SNN's validation is also limited to a single architecture and simple datasets.
Recently, STATA introduces an anchor token for token pruning with dual temporal and inter-layer alignment mechanisms, becoming the first token pruning method in spiking transformer validated on ImageNet \cite{zhuge2024towards}. However, it requires a complete retraining process, and the additional loss terms increase training overhead compared to uncompressed counterparts.

\section{Preliminary}

\paragraph{Spiking neuron model.}
Spiking neurons mimic the information transmission and processing of biological neurons. Due to the high computational complexity of biological neurons, researchers simplify spiking neurons into differential equations for computer simulation. 
The neural behavior of spiking neurons typically includes three mechanisms: membrane potential integration, spike generation, and reset \cite{wu2018spatio,neftci2019surrogate,zhang2021rectified}. Below, we describe these behaviors using the widely adopted Leaky Integrate-and-Fire (LIF) model, which can be described as follows:
\begin{align}
\label{eq:mem} 
\mathbf{\tilde u}^{\ell}[t]&=\mathbf{u}^{\ell}[t-1]+f({\mathbf{w}^{\ell}},\mathbf{s}^{\ell-1}[t]), &\text{(Voltage integration)},\\
\label{eq:lif}
\mathbf{s}^{\ell}[t]&=\mathrm{Heavside}(\mathbf{\tilde u}^{\ell}[t]-\theta),&\text{(Spike generation)},\\
\label{eq:headisde} 
\mathbf{u}^{\ell}[t] &= \begin{cases}
\mathbf{\tilde{u}}^{\ell}[t]\left(1- \mathbf{s}^{\ell}[t]\right), &\text{hard reset}, \\
\tau\mathbf{\tilde u}^{\ell}[t] - \theta\mathbf{s}^{\ell}[t], & \text{soft reset},
\end{cases} &\text{(Reset mechanism)},
\end{align}
where $\tau$ is the constant leaky factor, $t$ is the time step, $\mathbf{w}^{\ell}$ is the weight matrix of layer $\ell$, and $f(\cdot)$ is the convolution or linear operation followed by batch normalization (BN).
As described above, neurons integrate inputs and emit a spike $\mathbf{s}\in\{0,1\}$ when the membrane potential $\mathbf{\tilde{u}}$ exceeds the threshold $\theta$. 
After spike emission, the reset mechanism is invoked to update the membrane potential.

\paragraph{Spiking transformer.}

The spiking transformer architecture typically comprises four components: input embedding, spiking self-attention ($\op{SSA}$), multi-layer perceptron ($\op{MLP}$), and a classification head ($\op{CH}$). 
Given a 2D image sequence \( I \), the input embedding module linearly projects it into \( D \)-dimensional spiking features vector and partitions it into either a 2D grid of \( H \times W \) spiking patches or \( N \) flattened spiking patches. After adding positional encoding, the initial feature \(\mathbf{X}^0\) passes through \( L \) transformer blocks, each containing $\op{SSA}$ and $\op{MLP}$ modules. Finally, the features \(\mathbf{X}^L\) are aggregated via global average pooling ($\op{GAP}$) and processed by the $\op{CH}$ to generate predictions:
\begin{align}
\mathbf{X}^0&=\mathrm{InputEmbedding}\left(I\right), &&\mathbf X^0\in \mathbb{R}^{T \times H\times W\times D},  \\
\bf \hat{X}^\ell &= {\rm{SSA}}(\mathbf{X}^{\ell-1}) + \mathbf X^{\ell -1}, &&\mathbf X^\ell\in \mathbb{R}^{T \times H\times W\times D},\ell=1...L, \\
\bf X^\ell &= {\rm{MLP}}(\mathbf{\hat X}^\ell) + \mathbf{\hat{X}}^\ell, &&\mathbf{X^\ell}\in \mathbb{R}^{T \times H\times W\times D}, \ell=1...L, \\
\bf Y &= \op{CH}(\op{GAP}(\mathbf{X}^L)).
\end{align}
Notably, $\op{SSA}$ provides an efficient approach to model the local-global information of images using spike-based queries ($\mathbf{q}$), keys ($\mathbf{k}$), and values ($\mathbf{v}$) without employing softmax, described as, 
\begin{align}
&\rm SSA(\mathbf{X}^{\ell-1})=\mathcal{SN}((\mathbf{q}_{\bf s} \mathbf{k}_{\mathbf{s}}^\top) \mathbf{v}_{\bf s}),\\
&\mathbf x\mathbf{_s}=\mathcal{SN}(\mathbf x),\quad \mathbf{x} = \mathbf{w}_{\mathbf x}\cdot{\mathcal{SN}}(\mathbf{X}^{\ell-1}), \quad \mathbf{x}\in\{\mathbf{q}, \mathbf{k}, \mathbf{v}\}.
\label{eq:spikeqkv}
\end{align}
This design combines the efficiency of SNNs with the modeling capabilities of Transformers, enabling effective processing of information with reduced computational overhead \cite{yao2023spike}.

\section{Method}
In this section, we introduce our simple yet effective token pruning method for efficient spiking transformers. We first present the heuristic spatiotemporal information-retaining criterion that assesses token importance. Then, we introduce the information-retention pruning architecture with the block-level early stopping strategy. By integrating these two components, our TP-Spikformer achieves improved efficiency, scalability, and effectiveness.

\subsection{Heuristic spatiotemporal information-retaining criterion}
Extensive neuroscience research has shown that the human visual system does not process all information equally, but instead prioritizes regions that are spatially salient or exhibit significant temporal changes \cite{itti2002model,fecteau2006salience}. This selective mechanism allows biological systems to allocate computational resources efficiently to informative regions of visual input \cite{koch1987shifts}.
Inspired by this, we propose the IRToP to guide token pruning in spiking transformers, which assesses tokens based on their information content, giving tokens with richer spatial and temporal information higher preservation priority.

\paragraph{Spatial token scorer.}

In the human visual system, spatial locations compete for saliency within feature maps, allowing only those that quite differ from their local surroundings to persist for further processing \cite{itti2002model}. This motivates us to assess the representational divergence between each token and its spatial neighbors, assigning higher retention scores to those with greater spatial saliency.

In the spiking transformer framework, feature representations are structured either as spatial feature maps $\mathbf{X}^{\ell-1} \in \mathbb{R}^{T \times H \times W \times D}$ or in their flattened form $\mathbf{X}^{\ell-1} \in \mathbb{R}^{T \times N \times D}$, where $N = H \times W$ is the total number of tokens.
Consider a spatial feature map at time step $t$, for each token located at spatial position $(h, w)$, i.e., $\mathbf{X}^{\ell-1}_{t,h,w}\in \mathbb{R}^{D}$, we compute the dissimilarity between this token and a representative one in its spatial window.
The spatial dissimilarity of a single token is computed as,
\vspace{-3mm}
\begin{align}
&\mathcal{S}_\mathrm{score}(\mathbf{X}^{\ell-1}_{t,h,w}) = 1 -  \frac{\langle\mathbf{X}^{\ell-1}_{t,h,w}, \mathbf{Y}^{\ell-1}_{t,h,w}\rangle}{|\mathbf{X}^{\ell-1}_{t,h,w}| \cdot |\mathbf{Y}^{\ell-1}_{t,h,w}|},\quad \mathbf{Y}^{\ell-1}_{t,h,w}=\frac{1}{|\mathcal{W}_{h,w}|} \sum_{{(p,q)} \in \mathcal{W}_{h,w}}\mathbf{X}^{\ell-1}_{t,p,q},
\end{align}
where the representative token $\mathbf{Y}^{\ell-1}_{t,h,w}$ is the mean representation within the spatial window, capturing the local contextual information around the target token. Unlike pairwise similarity calculations with neighbor tokens, using this representative token reduces computational complexity.
$\mathcal{W}_{h,w}$ is the set of valid neighboring positions within the $k \times k$ window centered at the coordinate $(h, w)$, defined as,
\begin{equation}
\mathcal{W}_{h,w} = \{(h \pm \lfloor \frac{k-1}{2} \rfloor, w \pm \lfloor \frac{k-1}{2} \rfloor)\} \cap [0, H-1] \times [0, W-1].    
\end{equation}
We calculate the dissimilarity for each token in the feature map at time step $t$ and normalize these values to obtain the spatial saliency score. Each score is constrained between 0 and 1, with the sum of all scores equal to 1. A higher score indicates greater spatial saliency of a token, and these tokens are given higher priority for retention in token pruning. This ensures that the most spatially informative tokens of the original feature maps are preserved as much as possible during token pruning. Notably, this approach can be easily extended to frameworks with flattened token representations.

\begin{figure}[t]
  \centering
  \includegraphics[scale=0.47]{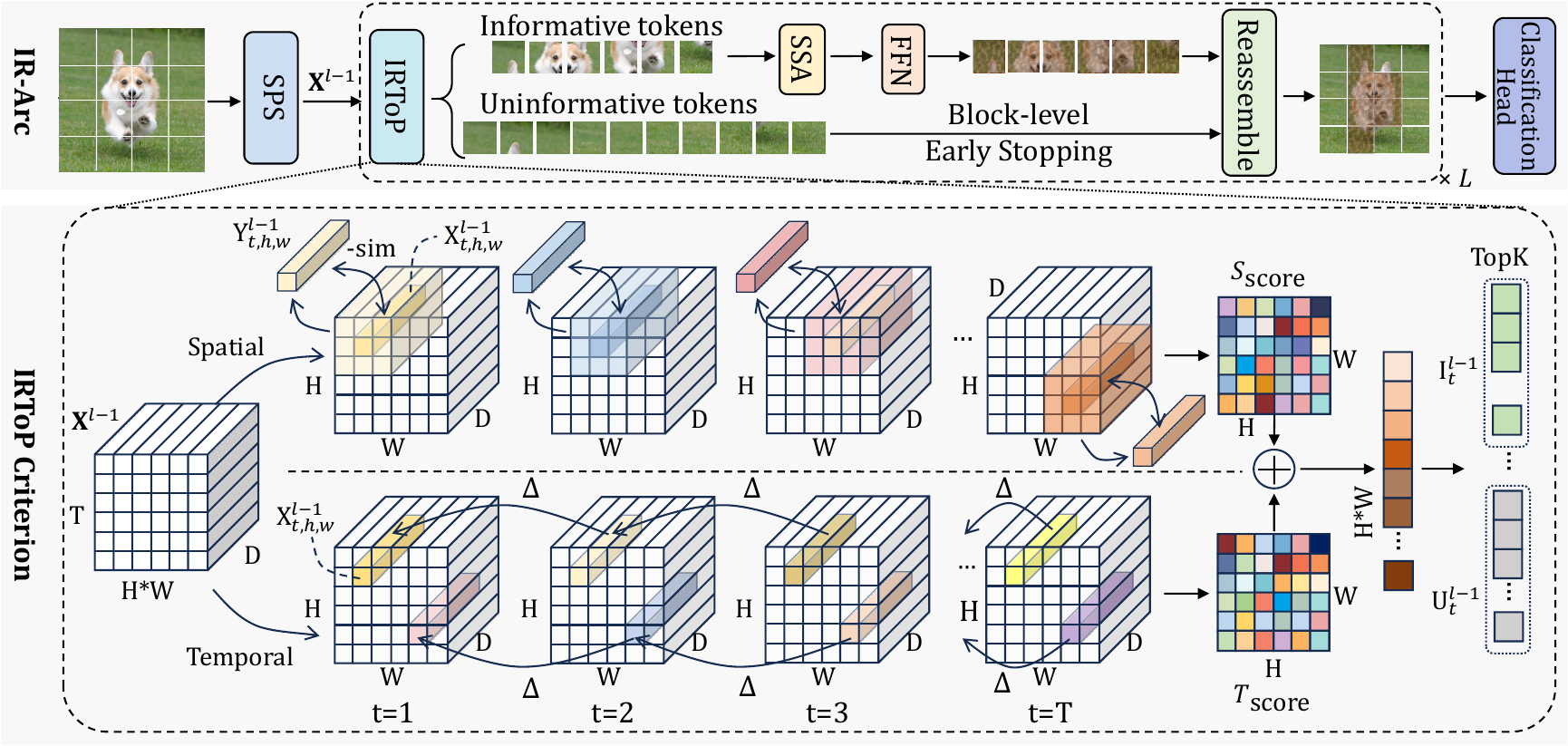}
  \caption{The overall workflow of the proposed TP-Spikformer, including the information-retention token pruning framework (\textbf{top}) and the spatiotemporal information-retaining criterion (\textbf{bottom}).}
  \label{fig:framework}
\end{figure}

\paragraph{Temporal token scorer.}
Neuroscientific research has demonstrated that the human visual system is highly sensitive to sudden and significant temporal changes \cite{rensink2002change,nothdurft2000salience}. This processing mechanism serves as an efficient information compression strategy, enabling the brain to swiftly locate and process key temporal dynamics within extensive visual input. Inspired by this, we measure the temporal dynamics of each token between consecutive time steps and assign higher retention scores to tokens exhibiting significant temporal variations. Considering a token at a position $(h, w)$ in the spatial feature map, we compute its temporal variation as,
\begin{equation}
\mathcal{T}_\mathrm{score}(\mathbf{X}^{\ell-1}_{t,h,w}) =
\begin{cases}
|\mathbf{X}^{\ell-1}_{t,h,w} - \mathbf{X}^{\ell-1}_{t-1,h,w}|, & \text{if } t > 1, \\
|\mathbf{X}^{\ell-1}_{t,h,w}|, & \text{if } t = 1.
\end{cases}
\end{equation}
We calculate and normalize the temporal variation of all tokens at time $t$ to obtain their temporal variation scores. Higher scores indicate richer temporal information of a token, leading to higher retention priority during pruning. This biologically inspired method allows spiking transformers to capture critical temporal features of input sequences while filtering out redundant information.

\paragraph{IRToP criterion.}
We combine the normalized spatial saliency and temporal variation scores for each token to obtain a spatiotemporal score at each time step, as detailed in Figure \ref{fig:framework}. Formally, for a token located at position $(h,w)$ at time step $t$, our IRToP criterion evaluates this token as follows,
\begin{equation}
\mathrm{IRToP}(\mathbf{X}^{\ell-1}_{t,h,w}) = \hat{\mathcal{S}}_\mathrm{score}(\mathbf{X}^{\ell-1}_{t,h,w}) + \hat{\mathcal{T}}_\mathrm{score}(\mathbf{X}^{\ell-1}_{t,h,w}),
\label{eq:irtop}
\end{equation}
where $\hat{\mathcal{S}}$ and $\hat{\mathcal{T}}$ denotes the normalized score. Given the $\ell$-th block's pruning rate $r_\ell$, we classify tokens in the input feature map $\mathbf{X}^{\ell-1}$ into informative and non-informative tokens based on their scores.
Specifically, we denote a set of \underline{t}oken \underline{s}cores in $\mathbf{X}^{\ell-1}$ at time step $t$ as $\mathrm{TS}_t^{\ell-1} = \{\mathrm{IRToP}(\mathbf{X}^{\ell-1}_{t,h,w})\}_{h=1,w=1}^{H,W}$, so the sets of informative tokens $\mathbf{I}_t^{\ell-1}$ and uninformative tokens $\mathbf{U}_t^{\ell-1}$ is:
\vspace{-5pt}
\begin{align}
&\mathbf{I}_t^{\ell-1} = \{\mathbf{X}^{\ell-1}_{t,h,w} \mid (h,w) \in \mathrm{TopK}(\mathrm{TS}_{t}^{\ell-1})\}, \quad \\
&\mathbf{U}_t^{\ell-1} = \{\mathbf{X}^{\ell-1}_{t,h,w} \mid \{(h,w)\}_{h=1,w=1}^{H,W} \notin \mathrm{TopK}(\mathrm{TS}_{t}^{\ell-1})\},
\end{align}
where $\mathrm{K} = \lceil(1-r_\ell) \!\times \!H \!\times\! W\rceil$ is the number of tokens to retain and $\mathrm{TopK}(\cdot)$ returns the highest-scoring token coordinates. The tokens in $\mathbf{U}_t^{\ell-1}$ are candidates for token pruning.
In summary, the IRToP criterion offers a neuroscience-inspired heuristic approach for token pruning in spiking transformers, enabling computational efficiency while retaining critical information.

\subsection{Information-retention token pruning architecture}
After categorizing informative and uninformative tokens based on the IRToP criterion, we propose an IR-Arc for token pruning in spiking transformers. The forward propagation of the $\ell$-th block at time step $t$ in our approach is described by the following formulas:
\begin{align}
& \mathbf{I}_t^{\ell-1}, \mathbf{U}_t^{\ell-1} \leftarrow  \{{\mathrm{IRToP}}(\mathbf{X}^{\ell-1}_{t,h,w})\}_{h=1,w=1}^{H,W}, && \mathbf{X}^{\ell-1}_t\in\mathbb{R}^{H\times W\times D},
\\
& {\mathbf{I'}}_t^{\ell} = {\rm{SSA}}(\mathbf{I}_t^{\ell-1}) + \mathbf{I}_t^{\ell-1}, &&\mathbf{I}_t^{\ell-1}\in \mathbb{R}^{ k\times D},\\
&\mathbf{X}_{t,\mathrm{inf}}^\ell = {\rm{MLP}}({\mathbf{I'}}_t^{\ell}) + {\mathbf{I'}}_t^{\ell}, &&\mathbf{I’}_t^{\ell-1}\in \mathbb{R}^{ k\times D},\\
&\mathbf{X}^{\ell}_{t,\mathrm{uni}}=\mathbf{U}_t^{\ell-1}, &&\mathbf{U}_t^{\ell-1}\in\mathbb{R}^{(H\times W-k)\times D},\\
&\mathbf{X}_t^\ell=\mathrm{Reassemble}(\mathbf{X}_{t,\mathrm{inf}}^\ell,\mathbf{X}^{\ell}_{t,\mathrm{uni}}), && \mathbf{X}_t^\ell \in \mathbb{R}^{H\times W \times D}.
\end{align}
In IR-Arc, informative tokens undergo complete SSA and MLP to further extract the essential features, while uninformative ones are skipped via block-level early stopping. All tokens are then reassembled into their original positions to restore the feature map size.
Unlike direct token removal, IR-Arc skips the calculation of uninformative tokens and then keeps them unchanged. This not only reduces memory and computational overhead, but also retains more information during token pruning. Additionally, the retention and reassembly strategy allows TP-Spikformer to be easily extended to hierarchical spiking transformers, like QKFormer \cite{zhou2024qkformer}, detailed in Appendix \ref{sec:why}.

We summarize the workflow of TP-Spikformer in Algorithm \ref{alg:ir_svit} and its advantages from two aspects. First, the information-retention strategy optimally allocates computational resources to tokens with high information content, enhancing efficiency without compromising model's feature extraction capabilities. Second, it reduces memory and computation cost by skipping uninformative tokens rather than removing them directly, ensuring compatibility with models with feature pyramids.
\vspace{-0.15cm}
\begin{algorithm}[h]
\setstretch{0.9}
\caption{The overall workflow of TP-Spikformer.}
\label{alg:ir_svit}
\begin{algorithmic}[0.85]
\small
\REQUIRE Trained spiking transformer model; Token pruning ratio per block $r=\{r_1,\cdots,r_L\}$; Input image $I$.
\ENSURE Classification results, with token pruning performed in forward propagation.
   \STATE $\triangleright$ $\mathbf{X}^0 \gets \mathrm{InputEmbedding}(I)$
   \FOR{$t \gets 1$ {\bfseries to} $T$}
   \FOR{$\ell \gets 1$ {\bfseries to} $L$}
   \STATE $\triangleright$ Get the input feature map: $\mathbf{X}^{\ell-1}_{t}$ and define an $\mathrm{avg\_kernel}$ with shape $[D,D,k,k]$ and value 1;
    \STATE $\triangleright$ Get representative tokens from each $k \times k$ spatial window: $\mathbf{Y}^{\ell-1}_t \gets$ Conv2d($\mathbf{X}^{\ell-1}_t$, $\mathrm{avg\_kernel}$);
    \STATE $\triangleright$ Token scoring: $\mathrm{TS}_t^{\ell-1} = \mathrm{SpatialScorer}(\mathbf{X}^{\ell-1}_{t}, \mathbf{Y}^{\ell-1}_{t}) + \mathrm{TemporalScorer}(\mathbf{X}^{\ell-1}_{t}, \mathbf{X}^{\ell-1}_{t-1})$;
        \STATE $\triangleright$ Select the $\rm K$ most informative tokens: $\mathbf{I}^{\ell-1}_{t} \gets \{\mathbf{X}^{\ell-1}_{t,h,w} \mid (h,w) \in \mathrm{TopK}(\mathrm{TS}^{\ell-1}_{t})\}$;
        \STATE $\triangleright$  Get the remaining uninformative tokens: $\mathbf{U}_t^{\ell-1} \gets \mathrm{All~Tokens} \setminus \mathrm{Informative~Tokens}$
        \STATE $\triangleright$ Extract and retain important information: $\mathbf{I'}^{\ell}_{t} \gets \mathrm{SSA}(\mathbf{I}^{\ell-1}_{t}) + \mathbf{I}^{\ell-1}_{t}$,~~$\mathbf{X}^{\ell}_{t,\mathrm{inf}} \gets \mathrm{MLP}(\mathbf{I'}^{\ell}_{t}) + \mathbf{I'}^{\ell}_{t}$;
        \STATE $\triangleright$ Early stopping for uninformative tokens: $\mathbf{X}^{\ell}_{t,\mathrm{uni}} \gets \mathbf{U}^{\ell-1}_{t}$;
        \STATE $\triangleright$ Reassemble tokens to restore the original feature map size: $\mathbf{X}^{\ell}_{t} \gets \mathrm{Reassemble}(\mathbf{X}^{\ell}_{t,\mathrm{inf}}, \mathbf{X}^{\ell}_{t,\mathrm{uni}})$;
       \ENDFOR
       \STATE $\triangleright$ $\mathrm{Y} = \op{CH}(\op{GAP}(\mathbf{X}^L_t))$;
   \ENDFOR
\end{algorithmic}
\end{algorithm}

\vspace{-0.3cm}
\section{Experiment}
\vspace{-0.1cm}

In this section, we conduct extensive experiments to assess our method. \textbf{First}, we evaluate TP-Spikformer's efficacy and efficiency on various architectures and tasks, comparing it with related work and uncompressed counterparts. \textbf{Second}, we study the zero-finetuning accuracy preservation property of TP-Spikformer. 
\textbf{Third}, we quantify the actual efficiency gains achieved by TP-Spikformer in training time and memory cost. \textbf{Finally}, we conduct ablation studies to validate the efficacy of IRToP and IR-Arc. We also visualize temporal and spatial scores of TP-Spikformer to provide insights into their operational mechanisms.
Appendix \ref{sec:appenExp} provides details about experimental setups.

\vspace{-0.15cm}
\subsection{Performance comparison}

\begin{table}[h]
\caption{Comparison of TP-Spikformer on small-scale datasets. `S' means the method doesn't add extra parameters or need retraining. $N_{\text{avg}}$ is the average token retention ratio. $^\dagger$ is an estimated token retention ratio based on reported metrics. Top three results are highlighted as \colorbox{color1}{first}, \colorbox{color2}{second}, and \colorbox{color3}{third}.}
\vspace{3mm}
\setlength{\tabcolsep}{3pt}
\label{tab:cifar}
\small
\begin{adjustbox}{max width=\linewidth}
\renewcommand\arraystretch{0.9}
\begin{tabular}{rcccccccccc}
\toprule 
\multirow{3}*{Method} & \multirow{3}*{S} & \multicolumn{3}{c}{CIFAR-10} & \multicolumn{3}{c}{CIFAR-100} & \multicolumn{3}{c}{DVS-CIFAR10} \\
\cmidrule(l{2pt}r{2pt}){3-5}\cmidrule(l{2pt}r{2pt}){6-8}\cmidrule(l{2pt}r{2pt}){9-11}
& &$T$  & $N_{avg}$ & {Acc. (\%)} &$T$ & $N_{avg}$ & {Acc. (\%)} &$T$ & $N_{avg}$ & {Acc. (\%)}\\
\midrule
Spikformer \cite{zhouspikformer} &-
 &  4 &  $\times 1$ &  95.19 \blutiny{Base} &  4 &  $\times 1$ &  78.21 \blutiny{Base} &  16 &  $\times 1$ &  80.9 \blutiny{Base}\\ 
\cdashline{1-11}[1pt/1pt]
\multirow{3}*{SparseSpikformer \cite{liu2024sparsespikformer}} &\xmark
& 4 & {$\times 0.85$} & \colorbox{color1}{95.18} \blutiny{(-0.01)}  & 4 & $\times 0.85$ & 77.70 \blutiny{(-0.51)} & 16 & $\times 0.85$ & {79.3} \blutiny{(-1.6)} \\ 
&\xmark & 4 & $\times 0.70$ & 95.03 \blutiny{(-0.16)}  & 4 & $\times 0.70$ & 77.07 \blutiny{(-1.14)} & 16 & $\times 0.70$ & 78.4 \blutiny{(-2.5)} \\
&\xmark & 4 & $\times 0.63$ & 94.77 \blutiny{(-0.42)}  & 4 & $\times 0.63$ & 76.78 \blutiny{(-1.43)} & 16 & \colorbox{color2}{$\times 0.63$} & 79.1 \blutiny{(-1.8)} \\
\cmidrule{2-11}
\multirow{2}*{AT-SNN \cite{kang2024snn}} &\xmark
& 4 & $\times 0.28$ & 95.06 \blutiny{(-0.13)} & 4 & $\times 0.75$ & \colorbox{color2}{78.14} \blutiny{(-0.07)} & - & - & -\\ 
&\xmark & 4 & \colorbox{color2}{$\times 0.21$} & 94.88 \blutiny{(-0.31)} & 4 & \colorbox{color3}{$\times 0.58$} & 77.27 \blutiny{(-0.94)} & - & - & -\\
\cmidrule{2-11}
STATA \cite{zhuge2024towards} &\xmark
& 4 &$\times$0.50$^\dagger$ & 95.00 \blutiny{(-0.19)} & 4 & \colorbox{color1}{$\times 0.50$}$^\dagger$ & 77.70 \blutiny{(-0.51)} & 16 & \colorbox{color1}{$\times 0.50$}$^\dagger$ &\colorbox{color2}{80.7} \blutiny{(-0.2)}\\
\midrule
\multirow{2}{*}{\textbf{TP-Spikformer}} &
\cmark & 4 & \colorbox{color3}{{$\times 0.25$}} & \colorbox{color2}{95.16} \blutiny{(-0.03)} & 4 & $\times 0.60$ & \colorbox{color1}{78.48} \blutiny{(+0.27)} & 16 & $\times 0.78$ & \colorbox{color1}{81.0} \blutiny{(+0.1)} \\ 
&\cmark & 4 & \colorbox{color1}{$\times 0.20$} & \colorbox{color3}{95.12} \blutiny{(-0.07)} & 4 & \colorbox{color2}{$\times 0.55$} & \colorbox{color3}{77.83} \blutiny{(-0.38)} & 16 & \colorbox{color1}{$\times 0.50$} & \colorbox{color2}{80.7} \blutiny{(-0.2)} \\   
\bottomrule
\end{tabular}
\end{adjustbox}
\end{table}

\begin{table}[h]
\vspace{-2mm}
\caption{Comparison of TP-Spikformer on ImageNet. `Thr' reports model throughput on one A800.}
\vspace{-3mm}
\label{tab:Imagenet}
\setlength{\tabcolsep}{1.5pt}
\renewcommand\arraystretch{0.93}
\begin{adjustbox}{max width=\linewidth}
\begin{tabular}{rrccccccc}
\toprule 
\makecell[c]{Method} & \makecell[c]{Architecture}  &S &$T$ & $N_{avg}$ & OPs$_{\mathrm{block}}$ (G) & Power (mJ) & Acc. (\%) & Thr (imgs/s) \\
\midrule
SEW \cite{fang2021deep} & SEW-ResNet-152 &-& 4 & $\times1$ &- & 12.891 & 69.26 &- \\
Spikformer \cite{zhouspikformer} & Spikformer-8-768 &-& 4 & $\times1$  &18.91 & 21.48 & 74.81 &229 \\
SNN-ViT \cite{wang2025spiking} & SNN-ViT-8-512 &- &4&$\times1$ & - &35.75&80.23&378\\
STATA \cite{zhuge2024towards} & Spikformer-8-768 &\xmark& 4 & $\times0.50^{\dagger}$  &-  & 11.16 & 74.03 &- \\
\midrule
\multirow{12}{*}{\textbf{TP-Spikformer}} 
&\multirow{4}*{\makecell[r]{SDT-V1-8-768 \\ \scriptsize{$\left [  \textcolor{black}{\textit{NeurIPS23}}\right ]$\cite{yao2023spike}}}} &- &4 & $\times 1$ &9.04 \blutiny{Base} &10.26 \blutiny{Base} &76.32 \blutiny{Base} &156 \blutiny{Base} \\ \cdashline{3-9}[1pt/1pt]
&&\cmark&4 &$\times0.74$ &6.75 \blutiny{($\downarrow$25\%)} &8.20 \blutiny{($\downarrow$19\%)} &75.82 \blutiny{(-0.50)} &181 \blutiny{($\uparrow$16\%)}\\
& &\cmark&4 &$\times0.65$ &5.93 \blutiny{($\downarrow$34\%)} &7.46 \blutiny{($\downarrow$26\%)} &75.62 \blutiny{(-0.70)} &189 \blutiny{($\uparrow$21\%)}\\
& &\cmark&4 &$\times0.51$ &4.71 \blutiny{($\downarrow$48\%)} &6.36 \blutiny{($\downarrow$38\%)} &74.79 \blutiny{(-1.53)} &202 \blutiny{($\uparrow$29\%)}\\
\cline{2-9}
&\multirow{4}*{\makecell[r]{QK-10-768 \\ \scriptsize{$\left [  \textcolor{black}{\textit{NeurIPS24}}\right ]$\cite{zhang2024qkformer}}}} &- & 4 & $\times1$ & 15.08 \blutiny{Base} & 32.12 \blutiny{Base} & 85.56 \blutiny{Base} &75 \blutiny{Base}\\ \cdashline{3-9}[1pt/1pt]
&&\cmark&4 &$\times0.72$  &10.7 \blutiny{($\downarrow$29\%)} &28.18 \blutiny{($\downarrow$12\%)} &84.45 \blutiny{(-1.11)} &84 \blutiny{($\uparrow$12\%)}\\
& &\cmark&4 &$\times0.65$  &9.61 \blutiny{($\downarrow$36\%)} &27.19 \blutiny{($\downarrow$15\%)} &84.32 \blutiny{(-1.24)} &88 \blutiny{($\uparrow$17\%)}\\
& &\cmark&4 &$\times0.53$  &7.97 \blutiny{($\downarrow$47\%)} &25.71 \blutiny{($\downarrow$20\%)} &82.53 \blutiny{(-3.03)} &106 \blutiny{($\uparrow$41\%)}\\
\cline{2-9}
&\multirow{4}*{\makecell[r]{SDT-V3-19M \\ \scriptsize{$\left [  \textcolor{black}{\textit{TPAMI25}}\right ]$}\cite{yao2025scaling}}} &- & 1$\times$4 & $\times1$ & 1.74 \blutiny{Base} & 5.47 \blutiny{Base} & 79.72 \blutiny{Base} &1562 \blutiny{Base}\\ \cdashline{3-9}[1pt/1pt]
& &\cmark&1$\times$4 &$\times0.78$ &1.37 \blutiny{($\downarrow$21\%)} &4.68 \blutiny{($\downarrow$14\%)} &79.01 \blutiny{(-0.71)}&1785 \blutiny{($\uparrow$14\%)}\\
& &\cmark&1$\times$4 &$\times0.65$ &1.13 \blutiny{($\downarrow$35\%)} & 4.43 \blutiny{($\downarrow$19\%)}&78.10 \blutiny{(-1.62)}&1851 \blutiny{($\uparrow$19\%)}\\
& &\cmark&1$\times$4 &$\times0.56$ &0.98 \blutiny{($\downarrow$44\%)} & 4.25 \blutiny{($\downarrow$22\%)}&77.55 \blutiny{(-2.17)}&1886 \blutiny{($\uparrow$21\%)}\\
\bottomrule
\end{tabular}
\end{adjustbox}
\vspace{-6mm}
\end{table}

\textbf{Image classification.}
We first compare TP-Spikformer with existing SNN token pruning methods. Existing methods are mostly validated on the Spikformer and small datasets, and we compare TP-Spikformer with them in Table \ref{tab:cifar} and \ref{tab:Imagenet}. Clearly, TP-Spikformer maintains high performance even under high compression ratios, e.g., retaining only 20\% tokens on CIFAR-10 with merely a 0.07\% accuracy drop.
Then, we assess TP-Spikformer on Imagenet-1K and various architectures, focusing on accuracy, block operations (OPs$_{\mathrm{block}}$), power, and throughput. As shown in Table \ref{tab:Imagenet}, reducing tokens greatly lowers OPs$_{\mathrm{block}}$ and power, with minimal accuracy loss. For instance, retaining only 53\% tokens in QKFormer cuts OPs$_{\mathrm{block}}$ by 47\%, power by 20\%, while maintaining 82.53\% accuracy.
This indicates TP-Spikformer serves as an effective and general token pruning method for SNNs.

\begin{figure}
\vspace{-4mm}
    \centering
    \begin{subfigure}[b]{0.255\textwidth}
        \centering
        \includegraphics[width=\textwidth]{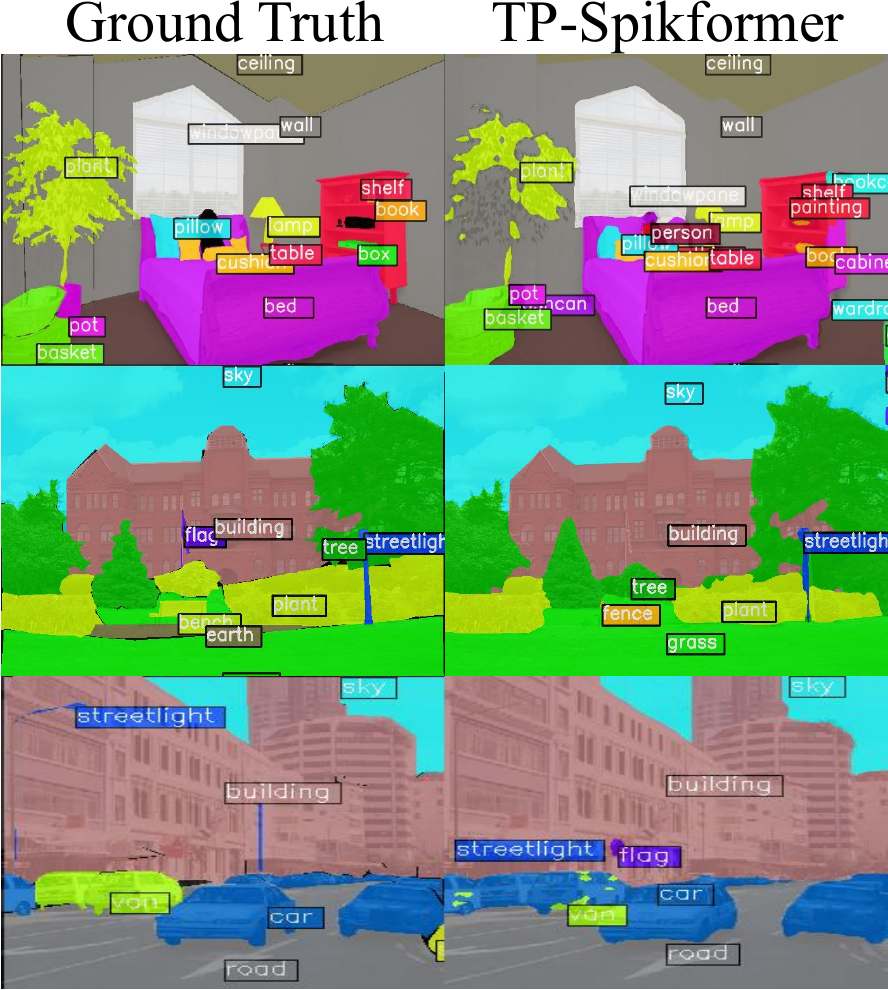}
        \caption{Semantic segmentation}
        \label{fig:seg}
    \end{subfigure}
    \hspace{-0.85mm}
    \begin{subfigure}[b]{0.255\textwidth}
        \centering
        \includegraphics[width=\textwidth]{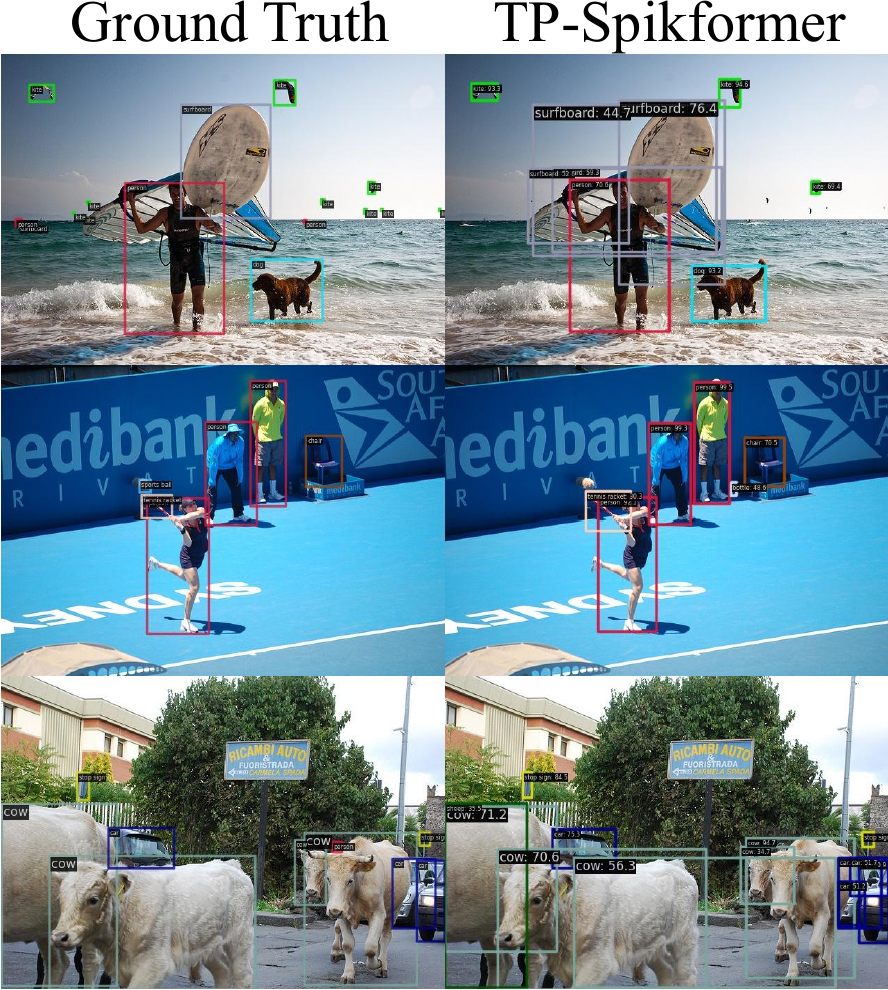}
        \caption{Object detection}
        \label{fig:det}
    \end{subfigure}
    \hspace{-0.85mm}
    \begin{subfigure}[b]{0.47\textwidth}
        \centering
        \includegraphics[width=\textwidth]{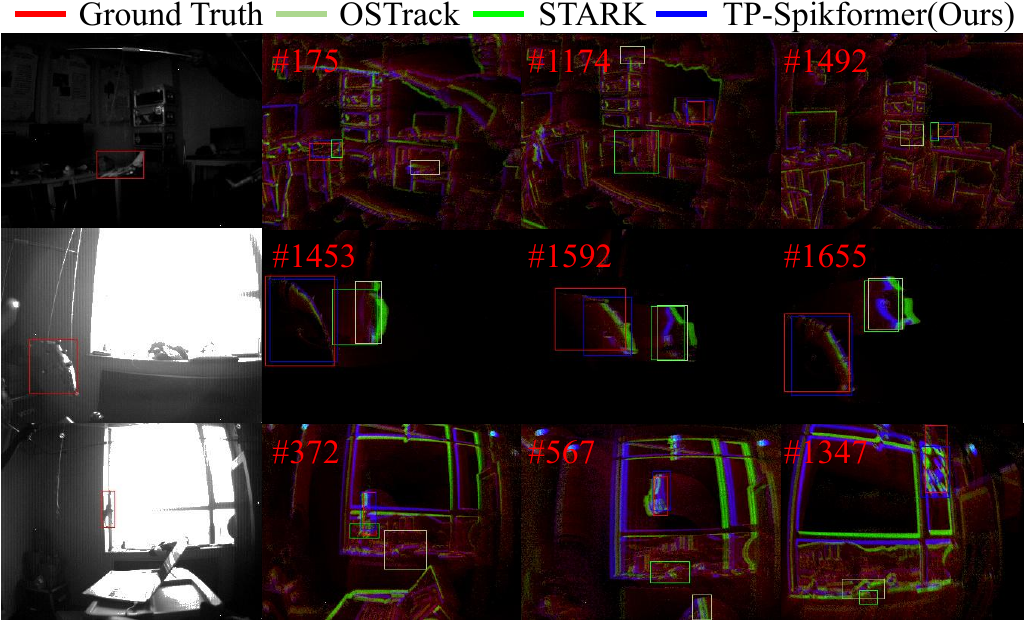}
        \caption{Event-based object tracking}
        \label{fig:sot}
    \end{subfigure}
    \vspace{-2mm}
    \caption{Visualization of ground truth and ours, showing its efficacy on diverse downstream tasks.}
    \label{fig:tasks}
\end{figure}

\textbf{Semantic segmentation.}
We use ADE20K to assess the efficacy of TP-Spikformer in semantic segmentation tasks. \textit{This comparison aims to show TP-Spikformer's competitiveness despite token pruning, not its superiority in performance.} We use TP-Spikformer with SDT-V3 and $N_{avg}$ of 0.78 and 0.56 as the backbone for feature extraction, and other settings follow \cite{yao2025scaling}. Results and visualizations are shown in Table~\ref{tab:seg} and Figure~\ref{fig:seg}. With only 56\% tokens retained, TP-Spikformer achieves a 1.7$\times$ throughput with only a 0.2\% mIoU drop compared to the uncompressed SDT-V3.

\begin{minipage}[c]{0.5\textwidth}
\begin{table}[H]
\centering
\small
\setlength{\tabcolsep}{1.2pt}
\caption{Segmentation result on ADE20K.}
\renewcommand\arraystretch{0.985}
\begin{tabular}{rcccccccc}
\toprule
\makecell[c]{Method} & $N_{avg}$ & \tabincell{c}{$T$}& \begin{tabular}[c]{@{}c@{}}Param  \\    (M)\end{tabular} &  \begin{tabular}[c]{@{}c@{}}Thr   \\   (imgs/s)\end{tabular}  &  \tabincell{c}{MIoU\\(\%)} \\
\midrule
\multirow{3}{*}{\begin{tabular}[c]{@{}c@{}}\citep{yaospike}\end{tabular}} & $\times 1$& 4 & 16.5 & 59.6  & 33.6    \\
& $\times 1$ & 1 & 58.9 & 37.6  & 34.8    \\
& $\times 1$ & 4 & 58.9 & 36.8  & 35.3    \\
\cmidrule{2-6}
\multirow{3}{*}{\begin{tabular}[c]{@{}c@{}}\citep{yao2025scaling}\end{tabular}}
& $\times 1$ & 2 & 11.0 & \colorbox{color3}{82.7}  & 31.9    \\
& $\times 1$ & 4 & 11.0 & 74.5  & \colorbox{color3}{40.1}    \\
& $\times 1$ & 8 & 11.0 & 70.4  & \colorbox{color1}{41.4}    \\
\midrule
\textbf{TP-Spikformer}&\colorbox{color2}{$\times0.78$}  & 4   & 11.0 & \colorbox{color2}{112.2}  & \colorbox{color2}{40.2}   \\
\textbf{TP-Spikformer}&\colorbox{color1}{$\times0.56$}  & 4   & 11.0 & \colorbox{color1}{128.6}  & {40.0}   \\
\bottomrule
\end{tabular}
\label{tab:seg}
\end{table}
\vspace{-2mm}
\end{minipage}
\hspace{-0.3cm}
\begin{minipage}[c]{0.5\textwidth}
\begin{table}[H]
\centering
\caption{Object detection results on COCO 2017.}
\setlength{\tabcolsep}{1.2pt}
\renewcommand\arraystretch{1}
\small
\begin{tabular}{rccccc}
\toprule
\makecell[c]{Method}  &$N_{avg}$ &\tabincell{c}{$T$}& \begin{tabular}[c]{@{}c@{}}Param    \\  (M)\end{tabular} &  \begin{tabular}[c]{@{}c@{}}Thr   \\   (imgs/s)\end{tabular}  &  \tabincell{c}{mAP\\@0.5(\%)} \\
\midrule
\citep{kim2020spiking}  &- & 3500 & 10.2 &- & 25.7 \\ 
\citep{li2022spike}   &-& 512 & 17.1 & -& 45.3 \\
\cmidrule{1-6}
\citep{zhang2023direct} &-&4 & 11.3 & -  & 28.5     \\
\citep{su2023deep}&-& 4   & 26.9 & -  & 50.1     \\
\multirow{1}{*}{\begin{tabular}[c]{@{}c@{}} \citep{yaospike}\end{tabular}} &$\times1$& 1 & 34.9 & \colorbox{color3}{32.6}  &{44.0}    \\
\multirow{1}{*}{\begin{tabular}[c]{@{}c@{}}\citep{yao2025scaling}\end{tabular}}&$\times1$
& 8 & 38.7 & 29.8  &  \colorbox{color1}{58.8}    \\
\midrule
\textbf{TP-Spikformer} &\colorbox{color2}{$\times0.78$} & 4   & 38.7 &  \colorbox{color2}{42.9}   & \colorbox{color2}{55.4}   \\
\textbf{TP-Spikformer} &\colorbox{color1}{$\times0.56$} & 4   & 38.7 &  \colorbox{color1}{43.6}   & \colorbox{color3}{54.4}  \\
\bottomrule
\end{tabular}
\label{tab:det}
\end{table}
\vspace{-2mm}
\end{minipage}

\textbf{Object detection.}
We use COCO2017 to evaluate the efficacy of TP-Spikformer, which also aims to show its competitiveness under token pruning instead of superior mAP.
We also use SDT-V3 with $N_{avg}$ of 0.78 and 0.56 as the backbone, and others follow \cite{yao2025scaling}. Results and visualizations are shown in Table~\ref{tab:det} and Figure~\ref{fig:det}. With only 78\% tokens and fewer time steps, TP-Spikformer reaches a 1.4$\times$ throughput with only a 1\% mAP drop, showing its efficacy in object detection.

\textbf{Event-based tracking.}
We select the event-based tracking task to verify the effect of TP-Spikformer in sequence vision tasks. Experiments are conducted on three benchmarks, i.e., FE108 \cite{zhang2021object}, FELT \cite{wang2024long}, and VisEvent \cite{wang2023visevent}.
Similar in segmentation and detection, we use SDT-V3 with $N_{avg}$ of 0.78 and 0.56 as the backbone, and other settings follow \cite{shan2025sdtrack}. Results and visualizations are shown in Table~\ref{tab:seg} and Figure~\ref{fig:sot}. Using only 56\% of the tokens, TP-Spikformer surpasses most RGB-based trackers and rivals the advanced SDTrack, demonstrating its effectiveness in sequential vision tasks.

\begin{table*}[h]
\centering
\small
\vspace{-2mm}
\setlength{\tabcolsep}{2pt}
\renewcommand\arraystretch{0.98}
\caption{TP-Spikformer vs. advanced trackers on three event-based object tracking benchmarks.}
\vspace{-2mm}
\begin{tabular}{@{}rccccccccc@{}}
\toprule
\multirow{3}{*}{\makecell[c]{Methods}} & \multirow{3}{*}{\begin{tabular}[c]{@{}c@{}}Time \\step \end{tabular}} &\multirow{3}{*}{$N_{avg}$} & \multirow{3}{*}{\begin{tabular}[c]{@{}c@{}}Power\\ (mJ)\end{tabular}} & \multicolumn{2}{c}{FE108} & \multicolumn{2}{c}{FELT} & \multicolumn{2}{c}{VisEvent} 
\\  \cmidrule(l){5-10} 
                              &                                               &       &                                                                & AUC(\%)         & PR(\%)          & AUC(\%)           & PR(\%)           & AUC(\%)         & PR(\%)         \\ \midrule
STARK~\cite{yan2021learning}                                      & 1     &$\times1$                & 58.88                                                                 & 57.4        & 89.2        & {\colorbox{color1}{39.6}}          & {\colorbox{color1}{51.7}}         & 34.1        & 46.8       \\
SimTrack~\cite{chen2022backbone}                                   & 1    &$\times1$                  & 93.84                                                                 & 56.7        & 88.3        & 36.8          & 47.0         & 34.6        & 47.6       \\
OSTrack\textsubscript{256}~\cite{ye2022joint}                                   & 1     &$\times1$                 & 98.90                                                                 & 54.6        & 87.1        & 35.9          & 45.5         & 32.7        & 46.4       \\
ARTrack\textsubscript{256}~\cite{wei2023autoregressive}                              & 1    &$\times1$                  & 174.8                                                                & 56.6        & 88.5        & \colorbox{color2}{39.5}          & 49.4         & 33.0        & 43.8       \\
SeqTrack-B\textsubscript{256}~\cite{chen2023seqtrack}                               & 1     &$\times1$                 & 302.7                                                                & 53.5        & 85.5        & 33.0          & 42.0         & 28.6        & 43.3       \\
HiT-B~\cite{kang2023exploring}                                     & 1       &$\times1$               & 19.78                                                                 & 55.9        & 88.5        & 38.5          & 48.9         & 34.6        & 47.6       \\
HIPTrack~\cite{cai2024hiptrack}                                 & 1        &$\times1$              & 307.7                                                                & 50.8        & 81.0        & 38.2          & 48.9         & 32.1        & 45.2       \\
ODTrack~\cite{zheng2024odtrack}                                & 1       &$\times1$               & 335.8                                                                & 43.2        & 69.7        & 29.7          & 35.9         & 24.7        & 34.7       \\
STNet~\cite{zhang2022spiking}                                     & 3      &$\times1$              & -                                                                     & -           & -           & -             & -            & 35.0        & \colorbox{color1}{50.3}       \\ 
{SDTrack$_{\rm Tiny}$ \cite{shan2025sdtrack}}   & 4        &$\times1$            &    {\colorbox{color3}{8.16}}                              &  \colorbox{color1}{59.0}       &     {\colorbox{color1}{91.3}}        &  {\colorbox{color3}{39.3}}   &  \colorbox{color2}{51.2}              &     {\colorbox{color1}{35.6}}      &   49.2      \\ 
\midrule
\textbf{TP-Spikformer}  &4  &\colorbox{color2}{$\times0.65$} &\colorbox{color2}{6.65} &\colorbox{color1}{59.0} &\colorbox{color2}{91.2}&39.1 &\colorbox{color3}{50.4} &\colorbox{color3}{35.3} &\colorbox{color2}{49.7} \\
\textbf{TP-Spikformer} &4 &\colorbox{color1}{$\times0.56$} &\colorbox{color1}{6.51} &\colorbox{color2}{58.4}&\colorbox{color3}{90.6}&38.9&50.0&\colorbox{color2}{35.2}&\colorbox{color3}{49.4}\\
\bottomrule
\end{tabular}
\label{main_table}
\end{table*}

\begin{figure}[H]
\vspace{-4mm}
    \centering
    \includegraphics[width=1\linewidth]{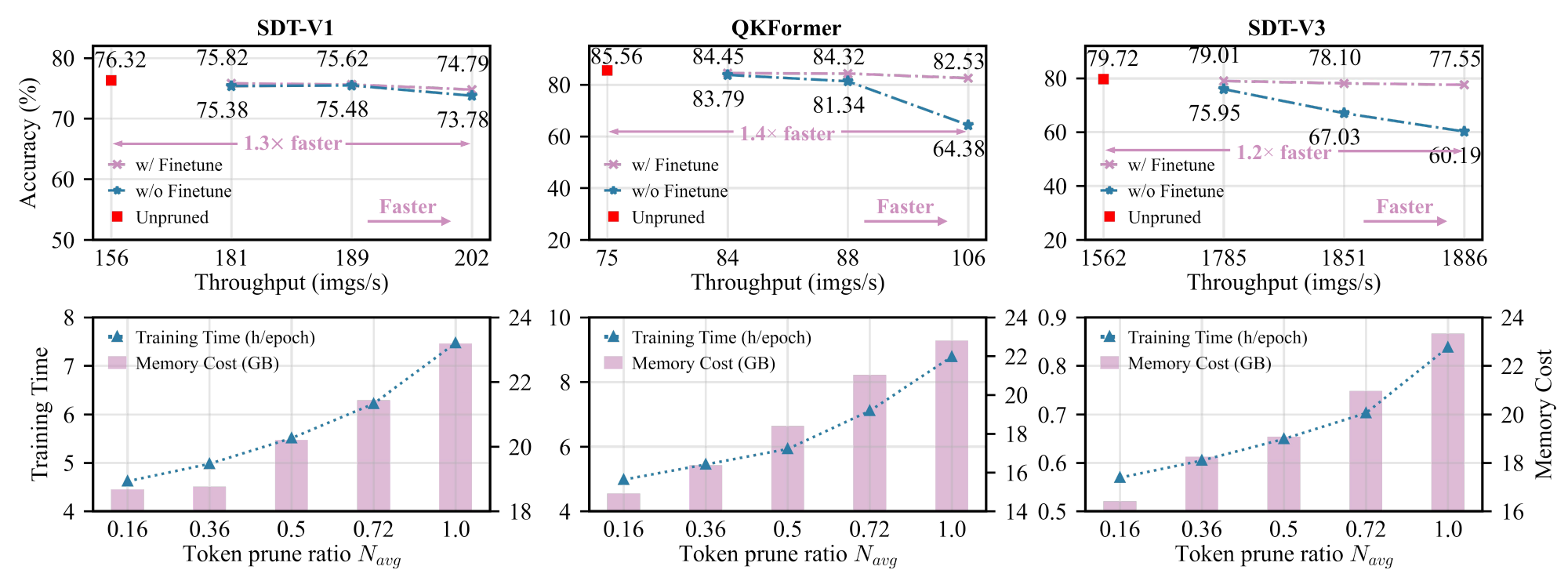}
    \vspace{-6mm}
    \caption{Zero-finetuning accuracy preservation (top) and efficiency gains (bottom) of TP-Spikformer.}
    \label{fig:comparisons}
\end{figure}

\vspace{-0.4cm}
\subsection{Validation and ablation study}
\label{sec:valstu}

\textbf{Zero-finetuning accuracy preservation of TP-Spikformer.}
We observe that TP-Spikformer can also perform well in a training-free manner. To verify this, we prune three architectures using official pre-trained weights, comparing results with and without fine-tuning. As shown in Figure \ref{fig:comparisons}(a), TP-Spikformer attains high accuracy without fine-tuning, showing its simplicity and generalization. This makes it well-suited for real-world scenarios with limited resources and no retraining budget.

\textbf{Speedup and memory improvement of TP-Spikformer.}
Besides inference throughput, we quantify the efficiency gains of TP-Spikformer in training. Experiments are conducted on ImageNet, measuring training time and memory usage. These metrics are tested on a single NVIDIA 4090, with the batch sizes of SDT-V1, QKFormer, and SDT-V3 set to 20, 15, and 200. As shown in Figure \ref{fig:comparisons}(b), TP-Spikformer notably reduces both training and memory cost as the token retention ratio decreases.

\textbf{Ablation study of IRToP and IR-Arc.}
We conduct ablation studies on ImageNet without fine-tuning, assessing IRToP and IR-Arc. Table \ref{tab:ab_acc} summarizes results, with key findings outlined below.
\textbf{First}, \texttt{IRToP} outperforms \texttt{Random} token pruning under \texttt{IR-Arc}, with gains of 13.76\%, 6.71\%, and 2.8\% on SDT-V1, QKFormer, and SDT-V3, respectively, showing its efficacy.
\textbf{Second}, the efficacy of \texttt{IR-Arc} is shown by comparing it with \texttt{Drop}; though their gap is small on SDT-V1 (59.88\% vs. 60.02\%), \texttt{IR-Arc} better supports varying feature map sizes, like QKFormer.
\textbf{Third}, decoupled analyses of \texttt{Temporal} and \texttt{Spatial} show that the \texttt{Spatial} scorer suffices on SDT-V1, while \texttt{Temporal} is more important on QKFormer, showing that both scorers in IRToP are indispensable.

\begin{table*}[t]
\centering
\caption{Ablation study. \texttt{Random} denotes random token pruning; \texttt{Drop} means token removal that reduces feature map size; \texttt{Spatial} and \texttt{Temporal} is using one single scorer for token selection.}
\vspace{-3mm}
\label{tab:ab_acc}
\setlength{\tabcolsep}{0pt}
\begin{adjustbox}{max width=\linewidth} 
\begin{tabular}{rccccc}
\toprule 
\makecell[c]{Architecture} & \texttt{[Random, Drop]} & \texttt{[Random, IR-Arc]}& \texttt{[Spatial, IR-Arc]}&\texttt{[Temporal, IR-Arc]}& \texttt{[IRToP, IR-Arc]}   \\
\midrule
SDT-V1$_{\times0.52}$& 59.88\%    & 60.02\%  & 73.52\%  & 70.95\% &73.78\% \\ 
QKFormer$_{\times0.65}$&  \texttt{\textcolor{black}{Fail}}   & 74.45\%  & 58.93\% &79.69\%  & 81.16\%\\ 
SDT-V3$_{\times0.78}$& \texttt{\textcolor{black}{Fail}}  & 73.15\%  & 75.95\%  & -&75.95\%  \\ 
\bottomrule
\end{tabular}
\end{adjustbox}
\end{table*}

\begin{figure}[h]
    \vspace{-2mm}
    \centering
    \includegraphics[width=1\linewidth]{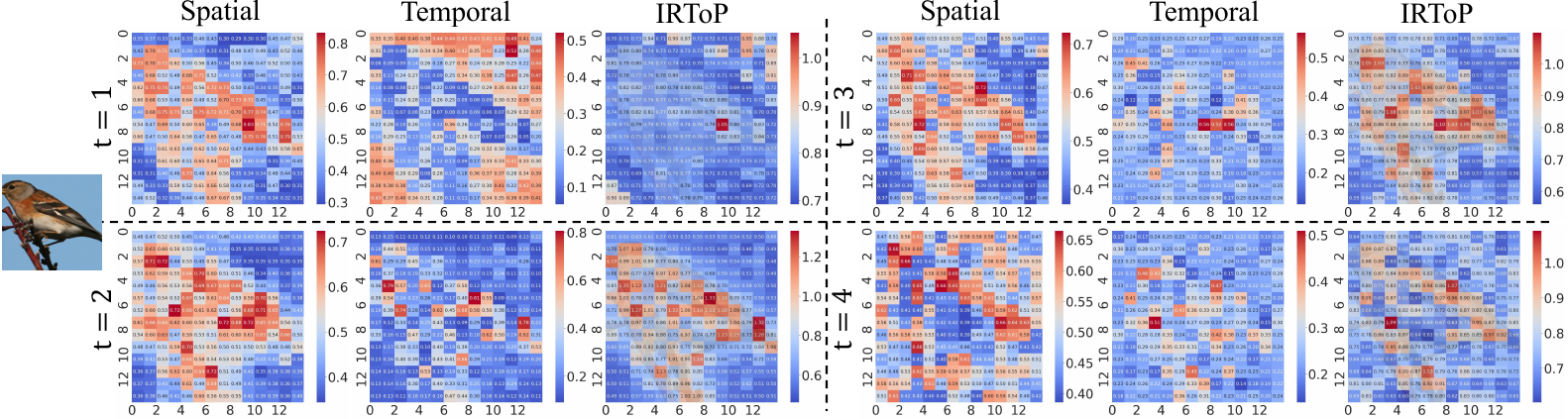}
    \caption{Visualization of spatial and temporal token scores in the 8th block of SDT-V1-8-768.}
    \label{fig:scorevis}
    \vspace{-2mm}
\end{figure}

\textbf{Decoupling analysis of IRToP.} 
We decouple and visualize spatial and temporal scores to understand their roles. Figure \ref{fig:scorevis} shows token scores from the last block before the classification head, where the spatial scorer assigns higher scores to tokens related to the main subject, while the temporal one emphasizes edges and key parts (e.g., claws, wings, beak). We find the temporal scorer underperforms at time step 1, likely due to the large magnitudes of background tokens causing misidentification. This also explains the poor token pruning results at the first time step in Figure \ref{fig:tokendrop}. We thus recommend using only spatial scores at the first step and combining both from the second step onward.

\section{Conclusion}
Existing transformer-based SNNs integrate transformer performance with SNN efficiency, yet are constrained by increased model size and computational demands. This paper presents TP-Spikformer, a simple yet effective token pruning approach for spiking transformers that reduces memory and computation overhead. Drawing inspiration from human visual processing, TP-Spikformer implements the IRToP criterion and IR-Arc architecture, striking an excellent balance between efficiency and performance across multiple architectures and tasks. Extensive experiments and comprehensive studies demonstrate its value in real-world scenarios with limited resources and no retraining budget.

\section*{Acknowledgments}
This work was supported by the National Natural Science Foundation of China (Grants 62576080 and 62220106008), the Sichuan Science and Technology Program (Grant 2024NSFTD0034), the Guangdong Introducing Innovative and Entrepreneurial Teams (Grant 2023ZT10×044), and the Shenzhen Science and Technology Research Fund (Grant JCYJ20220818103001002).

\bibliography{ref}
\bibliographystyle{iclr2026_conference}
\newpage
\appendix
\section{Experiment Details}
\label{sec:appenExp}
\subsection{Image Classification}
\paragraph{Dataset}
We evaluate TP-Spikformer on both static and dynamic datasets. The static datasets include CIFAR-10, CIFAR-100, and the large-scale ImageNet-1K. CIFAR-10 is a widely used benchmark in computer vision, containing 10 categories with 6,000 32$\times$32 images per category \cite{krizhevsky2009learning}. CIFAR-100 maintains the same image size but expands the categories to 100, which are organized into 20 superclasses \cite{krizhevsky2009learning}. ImageNet-1K is a large-scale vision dataset, featuring approximately 1.28 million training images and 50,000 test images across 1,000 categories \cite{deng2009imagenet}. Its diverse categories and rich image content make it a critical benchmark for image classification. Additionally, DVS-CIFAR10 is a dynamic dataset derived from CIFAR-10 using a dynamic vision sensor \cite{li2017cifar10}. This dataset includes 9,000 training samples and 1,000 test samples, with temporal resolution in the microsecond range and spatial resolution of 128$\times$128.

\begin{wraptable}{r}{0.5\textwidth}
\centering
\vspace{-0.5cm}
\setlength{\tabcolsep}{2pt}
\renewcommand{\arraystretch}{1.35}
\caption{Token preserving ratio on small datasets.}
\label{tab-cifar-rat}
\begin{tabular}{c|cc|cc|cc}
\hline
\multirow{2}{*}{\begin{tabular}[c]{@{}c@{}}Block\end{tabular}} & \multicolumn{2}{c|}{CIFAR-10}     & \multicolumn{2}{c|}{CIFAR-100}    & \multicolumn{2}{c}{DVS-CIFAR10}  \\ \cline{2-7} 
& \multicolumn{1}{c|}{\textbf{0.25}} & \textbf{0.20} & \multicolumn{1}{c|}{\textbf{0.60}} & \textbf{0.55} & \multicolumn{1}{c|}{\textbf{0.78}} & \multicolumn{1}{c}{\textbf{0.50}} \\ \hline
1     & \multicolumn{1}{c|}{0.76} & 0.56 & \multicolumn{1}{c|}{0.76} & 0.76 & \multicolumn{1}{c|}{0.78} & \multicolumn{1}{c}{0.76} \\ \hline
2     & \multicolumn{1}{c|}{0.14} & 0.14 & \multicolumn{1}{c|}{0.76} & 0.76 & \multicolumn{1}{c|}{0.78} & \multicolumn{1}{c}{0.25} \\ \hline
3     & \multicolumn{1}{c|}{0.06} & 0.06 & \multicolumn{1}{c|}{0.76} & 0.56 & \multicolumn{1}{c|}{-}    & \multicolumn{1}{c}{-}    \\ \hline
4     & \multicolumn{1}{c|}{0.06} & 0.06 & \multicolumn{1}{c|}{0.14} & 0.14 & \multicolumn{1}{c|}{-}    & \multicolumn{1}{c}{-}    \\
\hline
\end{tabular}
\end{wraptable}

\paragraph{Experimental setup}
For small-scale datasets CIFAR and DVS-CIFAR-10, we perform experiments with two token pruning ratio settings. Specifically, $N_{\text{avg}}$ is set to 0.25 and 0.20 for CIFAR-10, and 0.65 and 0.55 for CIFAR-100. We use the Spikformer-4-384 as used in studies \cite{liu2024sparsespikformer,kang2024snn,zhuge2024towards}, fine-tuning with a learning rate of 5e-5 and time step 4. For DVS-CIFAR10, we set $N_{\text{avg}}$ to 0.78 and 0.5, using the Spikformer-2-384 and fine-tuning with a learning rate of 7e-4.
Other experimental settings follow prior work.
For the large-scale ImageNet, we conduct experiments with three pruning ratios for each structure, detailed in Table \ref{tab:Imagenet}. We summarize the token retention ratio per block of small datasets in Table \ref{tab-cifar-rat} and Imagenet-1K in Table \ref{tab-rat}. In the next paragraph, we introduce how the token retention ratio per block is obtained. During fine-tuning, we remove the warm-up epoch and set the learning rate to 1e-5 for 50 epochs. The batch size per GPU for SDT-V1 and QKFormer is 72, while 760 for SDT-V3. The detailed settings for ImageNet are shown in Table~\ref{cls-set}.
In all experiments, when calculating the spatial scores of tokens, we compute the set of neighboring positions in a $3 \times 3$ window, i.e., $k=3$. We also discuss the effect of $k$ on performance in \ref{appensec:k}. 
\begin{table}[h]
\centering
\setlength{\tabcolsep}{9.2pt}
\renewcommand{\arraystretch}{1.2}
\caption{Token preserving ratio in each transformer block on Imagenet-1K.}
\label{tab-rat}
\begin{tabular}{c|ccc|ccc|ccc}
\hline
\multirow{2}{*}{\begin{tabular}[c]{@{}c@{}} Block\end{tabular}} & \multicolumn{3}{c|}{SDT-V1}                                  & \multicolumn{3}{c|}{QKFormer}                                & \multicolumn{3}{c}{SDT-V3}                                   \\ \cline{2-10} 
& \multicolumn{1}{c|}{\textbf{0.74}} & \multicolumn{1}{c|}{\textbf{0.65}} & \textbf{0.51} & \multicolumn{1}{c|}{\textbf{0.72}} & \multicolumn{1}{c|}{\textbf{0.65}} & \textbf{0.53} & \multicolumn{1}{c|}{\textbf{0.78}} & \multicolumn{1}{c|}{\textbf{0.65}} & \multicolumn{1}{c}{\textbf{0.56}} \\ \hline
1                                                                             & \multicolumn{1}{c|}{1}    & \multicolumn{1}{c|}{1}    & 0.73 & \multicolumn{1}{c|}{0.90} & \multicolumn{1}{c|}{0.81} & 0.64 & \multicolumn{1}{c|}{1} & \multicolumn{1}{c|}{0.81} & 0.64 \\ \hline
2                                                                             & \multicolumn{1}{c|}{1}    & \multicolumn{1}{c|}{0.73} & 0.73 & \multicolumn{1}{c|}{0.90} & \multicolumn{1}{c|}{0.81} & 0.64 & \multicolumn{1}{c|}{0.90} & \multicolumn{1}{c|}{0.81} & 0.64 \\ \hline
3                                                                             & \multicolumn{1}{c|}{0.73} & \multicolumn{1}{c|}{0.73} & 0.51 & \multicolumn{1}{c|}{0.72} & \multicolumn{1}{c|}{0.64} & 0.64 & \multicolumn{1}{c|}{0.90} & \multicolumn{1}{c|}{0.64} & 0.64 \\ \hline
4                                                                             & \multicolumn{1}{c|}{0.73} & \multicolumn{1}{c|}{0.73} & 0.51 & \multicolumn{1}{c|}{0.72} & \multicolumn{1}{c|}{0.64} & 0.49 & \multicolumn{1}{c|}{0.90} & \multicolumn{1}{c|}{0.64} & 0.64 \\ \hline
5                                                                             & \multicolumn{1}{c|}{0.73} & \multicolumn{1}{c|}{0.51} & 0.51 & \multicolumn{1}{c|}{0.72} & \multicolumn{1}{c|}{0.64} & 0.49 & \multicolumn{1}{c|}{0.64} & \multicolumn{1}{c|}{0.64} & 0.49 \\ \hline
6                                                                             & \multicolumn{1}{c|}{0.73} & \multicolumn{1}{c|}{0.51} & 0.51 & \multicolumn{1}{c|}{0.72} & \multicolumn{1}{c|}{0.64} & 0.49 & \multicolumn{1}{c|}{0.64} & \multicolumn{1}{c|}{0.56} & 0.49 \\ \hline
7                                                                             & \multicolumn{1}{c|}{0.51} & \multicolumn{1}{c|}{0.51} & 0.32 & \multicolumn{1}{c|}{0.64} & \multicolumn{1}{c|}{0.64} & 0.49 & \multicolumn{1}{c|}{0.64} & \multicolumn{1}{c|}{0.56} & 0.49 \\ \hline
8                                                                             & \multicolumn{1}{c|}{0.51} & \multicolumn{1}{c|}{0.51} & 0.32 & \multicolumn{1}{c|}{0.64} & \multicolumn{1}{c|}{0.56} & 0.49 & \multicolumn{1}{c|}{0.64} & \multicolumn{1}{c|}{0.56} & 0.49 \\ \hline
9                                                                             & \multicolumn{1}{c|}{-}    & \multicolumn{1}{c|}{-}    & -    & \multicolumn{1}{c|}{0.64} & \multicolumn{1}{c|}{0.56} & 0.49 & \multicolumn{1}{c|}{-}    & \multicolumn{1}{c|}{-}    & -    \\ \hline
10                                                                            & \multicolumn{1}{c|}{-}    & \multicolumn{1}{c|}{-}    & -    & \multicolumn{1}{c|}{0.64} & \multicolumn{1}{c|}{0.56} & 0.49 & \multicolumn{1}{c|}{-}    & \multicolumn{1}{c|}{-}    & -    \\ 
\hline
\end{tabular}
\end{table}

\begin{table}[ht]
\vspace{-0.4cm}
\centering
\setlength{\tabcolsep}{3pt}
\renewcommand{\arraystretch}{1.1}
\caption{Experimental setups on Imagenet-1K.}
\label{cls-set}
\adjustbox{}{}
\begin{tabular}{cccc}
\toprule
Hyper-parameter     & SDT-V1 & QKFormer & SDT-V3   \\ \midrule
$N_{\text{avg}}$    & 0.74, 0.65, 0.51 & 0.72, 0.65, 0.51 & 0.78, 0.65, 0.56   \\
$k$ in IRToP&3&3&3\\
Time step            & 4 & 4 & 4   \\
Warmup epoch        & None & None & None \\
Epoch              & 50 & 50 & 50               \\
Resolution          & 224$\times$224 & 224$\times$224 & 224$\times$224     \\
Batch size per GPU          & 72 & 72 & 760        \\
Optimizer           & Adam & Adam & Adam    \\
Weight decay        & 0           & 0      &0    \\
Initial learning rate  & 1e-5 & 1e-5 & 1e-5    \\
Learning rate decay & Cosine     & Cosine   & Cosine    \\
\bottomrule 
\end{tabular}
\end{table}

In order to find the optimal pruning combination between blocks, we employ a search strategy before fine-tuning to determine the token pruning ratio per block based on the given global pruning rate. The grid search used is a very simple method, which is intended to perform a coarse search to initially identify a reasonable pruning rate. Given a pre-trained model and a global token preservation, we summarize its detailed search process below.
\begin{itemize}
    \item First is to obtain a set of token preservation ratio combinations. Specifically, the search space for ratios is restricted to a small set of discrete values, such as 0.9×0.9, 0.8×0.8, 0.75×0.75, 0.6×0.6, etc. Furthermore, we impose a monotonic constraint, requiring the token preservation ratio to decrease progressively from shallow to deeper blocks. This is motivated by the observation that shallow layers capture low-level features and thus require higher token retention, while deeper layers handle high-level semantic information and can tolerate more aggressive token pruning \cite{lin2021channel}.
    \item Second, we randomly sample a small batch of data from the training dataset and evaluate the accuracy for each combination in the combinations set. The combination with the top-1 accuracy is selected and used for subsequent fine-tuning.
\end{itemize}
\textbf{By reviewing the search logs, we observe that different configurations give similar performance under the same global ratio. Therefore, the grid search is only used for SDT-V3, while for QKFormer and SDT-V1, we directly set the ratios manually and fine-tune the models without performing grid search.} We summarize the search details for SDT-V3 in Table \ref{tab:search}, including the discrete search space, the number of candidate combinations, the time to evaluate each combination, and the total search time. 

\begin{table}[htbp]
\centering
\caption{Search details for the SDT-V3.}
\setlength{\tabcolsep}{2pt}
\label{tab:search}
\begin{tabular}{@{}lcccccc@{}}
\toprule
Model & $N_{avg}$ & \makecell[c]{Searching space per\\ token retention ratio} & \makecell[c]{Number of \\combinations} & \makecell[c]{Evaluation time \\per combination} & \makecell[c]{Total Time \\(4*NVIDIA 4090)} \\
\midrule
SDT-V3 & 0.78 & [1,0.90,0.81,0.72,0.64,0.56,0.49] & 65 & 26s & 28min 11s \\
SDT-V3 & 0.65 & [0.81,0.72,0.64,0.56,0.49,0.42,0.36] & 89 & 22s & 32min 38s \\
SDT-V3 & 0.56 & [0.81,0.72,0.64,0.56,0.49,0.42,0.36] & 166 & 16s & 44min 16s \\
\bottomrule
\end{tabular}
\end{table}

\subsection{Semantic Segmentation}
\paragraph{Dataset}
ADE20K~\cite{zhou2019semantic} is a widely used and well-established dataset for semantic segmentation in computer vision research. It comprises approximately 25,000 images, with over 20,000 images designated for training, 2,000 images for validation, and 3,000 images for testing. Each image in the dataset is densely annotated with pixel-level labels across 150 distinct semantic categories. These categories cover a wide array of objects, such as people, cars, and animals, as well as scene elements like sky, roads, and vegetation, each with intricate visual features that make semantic segmentation tasks more challenging. Due to its diversity and complexity, ADE20K serves as a critical and challenging benchmark for evaluating the performance of segmentation algorithms. 

\paragraph{Experimental setup}
In this work, we begin by converting the \emph{mmsegmentation} \cite{contributors2020mmsegmentation} codebase to its spike-based version, inspired by the SDT-V3 \cite{yao2025scaling}. 
We employ TP-Spikformer with SDT-V3-19M as the backbone for feature extraction, integrated with spike FPN \citep{kirillov2019panoptic} for segmentation. 
The backbone is initialized using pretrained weights
from ImageNet, ensuring that the network has a strong starting point for feature extraction.
The newly added layers are initialized using the Xavier method \cite{glorot2010understanding}.
The experimental settings follow the parameters set in SDT-V3 to ensure consistency and comparability.
We fine-tune the model with two ratios same as object detection on 4$\times$4090 with a batch size of 12 per GPU, while original SDT-V3 is limited to 8. The results in Table~\ref{tab:seg} show that our method maintains the performance of SDT-V3 and greatly increases throughputs.
This comparison is not intended to demonstrate that our method achieves top-1 accuracy, but rather to highlight that our approach remains competitive even under token pruning conditions.

\subsection{Object Detection}
\paragraph{Dataset}

We evaluate TP-Spikformer on COCO2017~\cite{lin2014microsoft}, a large-scale benchmark that is widely used for object detection tasks. The dataset comprises a total of 118K training images, 5K validation images, and 40K test images, providing a comprehensive and diverse set of visual data. It covers 80 object categories, including everyday items such as cars, bicycles, animals, and household objects, which are essential for testing the algorithm’s ability to recognize and interpret a wide range of visual content. In addition to object categories, COCO offers multiple types of annotations, including object instance segmentation masks, keypoints, and captions, all of which contribute to the dataset’s robustness for evaluating various vision tasks. Notably, COCO emphasizes contextual relationships between objects within complex, everyday scenes, offering a more realistic and challenging evaluation setting compared to simpler datasets. This makes COCO a crucial benchmark for assessing the performance of computer vision algorithms in real-world, practical applications.

\paragraph{Experimental setup}
Similar to semantic segmentation, we begin by converting the \emph{mmdetection} \cite{mmdetection} codebase to its spike-based version. Our model architecture integrates TP-Spikformer with Mask R-CNN \citep{he2017mask}. The backbone is initialized using pretrained weights from ImageNet, and the newly added layers are initialized using the Xavier method \citep{glorot2010understanding}. We fine-tune the model with two different average token retention ratios: 0.56 and 0.78. These two settings allow us to explore how different levels of token retention influence the model's performance in object detection and segmentation tasks. The experiments are conducted on a 4$\times$A800 setup, with a batch size of 5 per GPU, providing ample computational resources to handle the large-scale training process.
The results are summarized in Table \ref{tab:det}, which also highlights that our approach remains competitive in object detection tasks.

\subsection{Event-based Tracking}
\paragraph{Dataset}
We use three event-based tracking benchmarks to assess our TP-Spifkormer, detailed as:
\begin{itemize}
    \item FE108 is captured by the DAVIS346 dynamic vision sensor, with an event rate spanning a range from 0 to 3800 events/ms \cite{zhang2021object}. This dataset features 21 diverse target categories. The diversity of categories and the high event rate make FE108 particularly useful for evaluating event-based models under varying conditions.
    \item FELT~\cite{wang2024long} is specifically designed to address the challenges associated with long-term object tracking in dynamic environments. This dataset places a strong emphasis on scenarios where the loss and recovery of targets are crucial for maintaining tracking accuracy. 
    \item VisEvent~\cite{wang2023visevent} is a large-scale dataset dedicated to event-based visual tasks, offering a robust testing ground for various event-driven models and algorithms under extreme conditions. With its broad scope, VisEvent includes a wide range of event-based visual tasks, providing a unique and challenging environment for assessing model performance. 
\end{itemize}
These datasets serve as three of the most important benchmarks in event-based tracking. Their diversity makes them important for evaluating the performance of event-driven models and algorithms.

\paragraph{Experimental setup}
We use the SDTrack pipeline to build a tracker for event-based tracking tasks \cite{shan2025sdtrack}. Specifically, we train the tracker using an image pair matching task \cite{chen2022backbone,yan2021learning,ye2022joint} and employ weighted focal loss \cite{law2018cornernet} for classification. For the predicted bounding boxes, L1 loss and generalized IoU loss \cite{rezatofighi2019generalized} are used for bounding box regression. We train the model for 100 epochs on the FE108 and VisEvent datasets, using a pretrained ImageNet-1K model, and for 300 epochs on the FELT dataset. For each training epoch on the FE108 and FELT datasets, we randomly sample 60,000 sample pairs with a maximum interval of 200, while on VisEvent, we use 30,000 pairs. The learning rate used during training is 4e-4, decaying to 4e-5 at 80\% of the training progress. We apply normalization and regularization on the FELT dataset, and a Hanning window penalty is used to constrain the predicted boxes. However, no data augmentation or preprocessing is applied to the FE108 and VisEvent. All of the above experimental settings are strictly aligned with SDTrack \cite{shan2025sdtrack}.

\section{Measurement of Efficiency Metrics}

\begin{table}[h]
\vspace{-0.3cm}
\centering
\setlength{\tabcolsep}{2pt}
\renewcommand{\arraystretch}{1.2}
\caption{Training time and memory usage of TP-Spikformer under different ratios and architectures.}
\label{tab:met}
\begin{adjustbox}{max width=\linewidth}
\begin{tabular}{c|cc|cc|cc}
\hline
\multirow{3}{*}{{$N_{\text{avg}}$}} & \multicolumn{2}{c|}{SDT-V1}    & \multicolumn{2}{c|}{QKFormer}      & \multicolumn{2}{c}{SDT-V3}       \\ \cline{2-7} 
                     & \multicolumn{1}{c}{{\begin{tabular}[c]{@{}c@{}} \small{Memory} \\ \small{usage (GB)} \end{tabular}}} & {\begin{tabular}[c]{@{}c@{}} \small{Training} \\ \small{time (h/epoch)} \end{tabular}} & \multicolumn{1}{c}{{\begin{tabular}[c]{@{}c@{}} \small{Memory} \\ \small{usage (GB)} \end{tabular}}} & {\begin{tabular}[c]{@{}c@{}} \small{Training} \\ \small{time (h/epoch)} \end{tabular}} & \multicolumn{1}{c}{{\begin{tabular}[c]{@{}c@{}} \small{Memory} \\ \small{usage (GB)} \end{tabular}}} & {\begin{tabular}[c]{@{}c@{}} \small{Training} \\ \small{time (min/epoch)} \end{tabular}} \\ \hline
$\times1$     & \multicolumn{1}{c}{23.19}  &  7.47 & \multicolumn{1}{c}{22.80}  & 8.78  & \multicolumn{1}{c}{23.33}  & 50.27  \\ 
$\times0.72$  & \multicolumn{1}{c}{21.44}  &  6.22 & \multicolumn{1}{c}{21.03}  & 7.11  & \multicolumn{1}{c}{20.96}  & 42.16  \\ 
$\times0.50$  & \multicolumn{1}{c}{20.21}  &  5.51 & \multicolumn{1}{c}{18.40}  & 5.93  & \multicolumn{1}{c}{19.08}  & 38.96  \\ 
$\times0.36$  & \multicolumn{1}{c}{18.76}  &  4.98 & \multicolumn{1}{c}{16.37}  & 5.45  & \multicolumn{1}{c}{18.25}  & 36.29  \\ 
$\times0.16$  & \multicolumn{1}{c}{18.68}  &  4.62 & \multicolumn{1}{c}{14.92}  & 4.98  & \multicolumn{1}{c}{16.42}  & 34.16  \\ 
\hline
\end{tabular}
\end{adjustbox}
\end{table}

As shown in Figure~\ref{fig:comparisons}, we report the training time per epoch, GPU memory usage, and inference throughput for SDT-V1, QKFormer, and SDT-V3. In this section, we provide a detailed description of how these metrics are measured, including the measurement method, experimental setup, and results.

The training time per epoch and GPU memory usage are measured by monitoring the time taken to complete one epoch and the GPU memory consumption during training. These metrics are tested on a single NVIDIA 4090, with batch sizes fixed at 20 for SDT-V1, 15 for QKFormer, and 200 for SDT-V3. We measure these two metrics of TP-Spikformer under different token pruning rates, with the results summarized in Table~\ref{tab:met}. It is evident that the proposed TP-Spikformer significantly reduces both training time and memory consumption, resulting in substantial efficiency gains. As a result, under the same configuration, when maximizing GPU resource utilization, TP-Spikformer typically allows for a higher batch size compared to its uncompressed counterpart. 

For the metric of inference throughput, we estimate it by calculating the number of images processed per second during the inference process. The throughput values reported in Figure \ref{fig:comparisons} and Tables~\ref{tab:Imagenet}-\ref{tab:det} are measured on a single NVIDIA A800 GPU, with a batch size of 36 for both SDT-V1 and QKFormer, and 1024 for SDT-V3 in the classification task. In the case of segmentation and detection tasks, the batch size is fixed to 1. The obtained throughput results, presented in Tables~\ref{tab:Imagenet} to \ref{tab:det}, display that TP-Spikformer achieves higher throughput across all models while maintaining high accuracy. This enhances the efficiency of model inference and faster real-time processing, making it more suitable for deployment in resource-constrained scenarios that require real-time processing.

\section{Analysis of $k$ in the IRToP criterion}
\label{appensec:k}
In this section, we evaluate the impact of different values of $k$ on the performance of TP-Spikformer, specifically SDT-V1, QKFormer, and SDT-V3, on the ImageNet-1K dataset under zero-finetuning conditions. The results are summarized in Table \ref{tab:k}. Each architecture shows stable performance across the different values of $k$, with a slight decrease in accuracy as $k$ increases. This drop in performance may be attributed to the fact that larger spatial windows, while capturing more global context, reduce focus on important local details, which is crucial for tasks like fine-grained detection and segmentation. Therefore, we select $k = 3$ for the experiments presented in the main text, as it offers a good balance between computational efficiency and effectiveness.

\begin{table}[h]
\vspace{-0.3cm}
\centering
\renewcommand{\arraystretch}{1.3}
\caption{Performance of TP-Spifkormer with different $k$ on ImageNet-1K without finetuning.}
\label{tab:k}
\begin{tabular}{c|ccc|ccc|ccc}
\hline
\multirow{2}{*}{$k$ value} & \multicolumn{3}{c|}{SDT-V1}                   & \multicolumn{3}{c|}{QKFormer}                 & \multicolumn{3}{c}{SDT-V3}                    \\ \cline{2-10} 
& \textbf{0.74} & \textbf{0.65} & \textbf{0.51} & \textbf{0.72} & \textbf{0.65} & \textbf{0.53} & \textbf{0.78} & \textbf{0.65} & \textbf{0.56} \\ \hline
3     & {75.38}         & {75.48}         & {73.78}         & {83.79}         & {81.34}         & {64.38}         & {75.95}         & {67.03}         & {60.19}         \\ \hline
5     & 75.37         & 75.48         & 73.77         & 83.65         & 81.23         & 64.18         & 75.77         & 66.59         & 59.71         \\ \hline
7     & 75.34         & 75.48         & 73.77         & 83.66         & 81.20         & 64.32         & 75.59         & 66.47         & 59.40         \\ \hline
\end{tabular}
\end{table}

\section{Adaptive weighting of spatial and temporal scorers in IRToP}
As for Eq. \ref{eq:irtop}, we have explored adaptive weighting between spatial and temporal components to analyze the effects of adaptive weight in the IRToP criterion. Specifically, we introduce a learnable parameter $\alpha$ to balance the spatial and temporal scores adaptively. The modified IRToP criterion is formulated as:
\begin{align}
    \mathrm{IRToP}(\mathbf{X}^{\ell-1}_{t,h,w}) = \alpha\times\hat{\mathcal{S}}_\mathrm{score}(\mathbf{X}^{\ell-1}_{t,h,w}) + (1-\alpha)\times\hat{\mathcal{T}}_\mathrm{score}(\mathbf{X}^{\ell-1}_{t,h,w}),
\end{align}
where $\alpha$ is initialized to 0.5 (equal weighting) and is differentiable, allowing it to be optimized during training. This enables the model to automatically learn the optimal balance between spatial and temporal importance. We conduct experiments on ImageNet-1K using SDT-V1 with a pruning ratio of 0.51, fine-tuning for 50 epochs. The results are summarized Table \ref{tab:adap}. Compared to fixed equal weighting ($\alpha$ = 0.5), we observe a slight performance drop when using the adaptive weighting method. We suspect this is because the learned value of $\alpha$ converges to around 0.3, suggesting that the model tends to emphasize temporal features over spatial ones. In this case, the model may overly focus on critical temporal dynamics and local details, potentially at the expense of broader spatial context that is essential for robust feature representation, leading to the observed performance degradation.
\begin{table}[htbp]
\centering
\setlength{\tabcolsep}{10pt}
\caption{Analysis of adaptive weighting of spatial and temporal scorers in IRToP.}
\label{tab:adap}
\begin{tabular}{@{}lcccc@{}}
\toprule
Model & Ratio & \makecell[c]{Final ratio $\alpha$} & \makecell[c]{Fine-tuning accuracy\\under adaptive $\alpha$}&\makecell[c]{Fine-tuning accuracy\\under fixed $\alpha$=0.5} \\
\midrule
SDT-V1 & 0.51 & 0.3 & 74.23\% & 74.79\% \\
\bottomrule
\end{tabular}
\end{table}

\section{Detailed Analysis of IRToP}
\begin{figure}[h]
    \centering
    \includegraphics[width=1\linewidth]{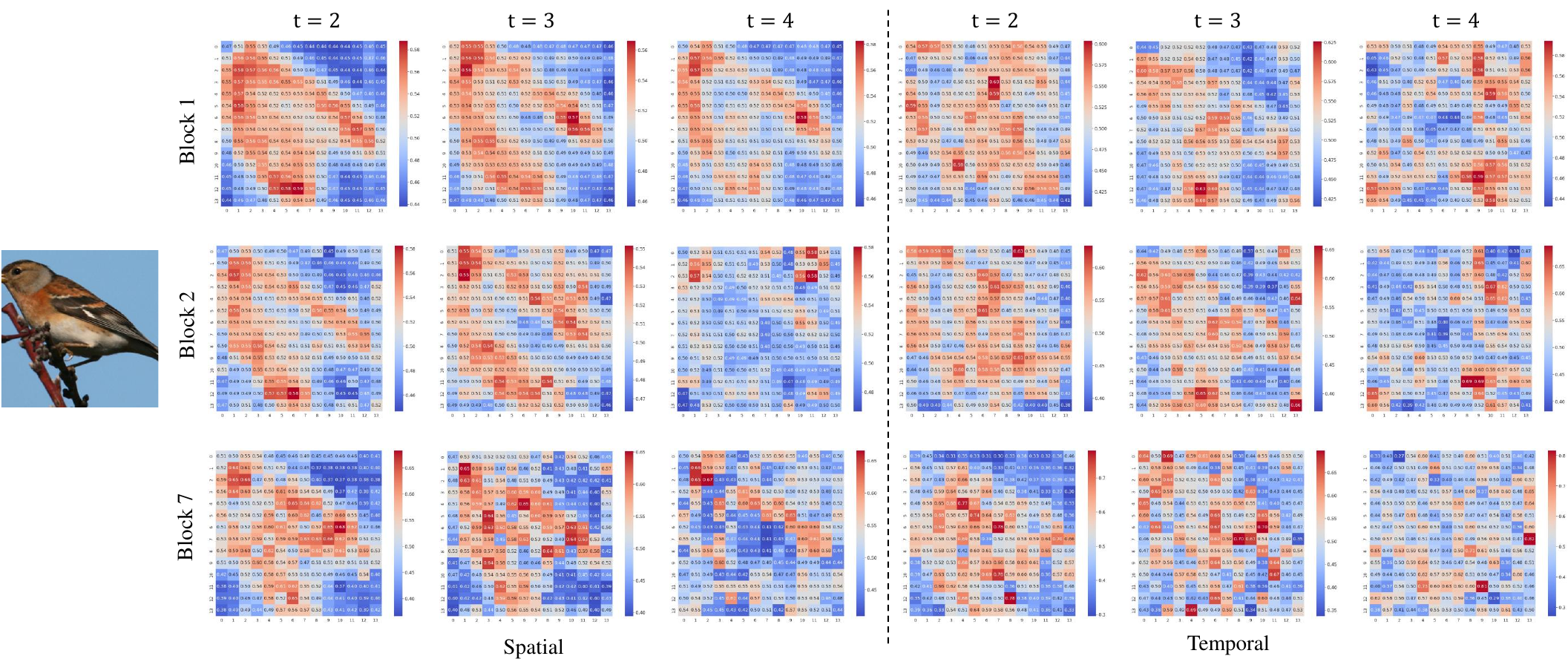}
    \caption{Detailed visualization of spatial and temporal token scores of SDT-V1-8-768.}
    \label{fig:irtop}
\end{figure}

IRToP proves effective by selecting informative tokens with two key characteristics: tokens that represent the overall outline and tokens that capture specific detail features. As a supplement to Figure \ref{fig:scorevis}, we further analyze the scores of tokens selected from the 1st, 2nd, and 7th encoder blocks. In Figure \ref{fig:irtop}, we decouple and visualize their spatial and temporal scores.
From the spatial scores, it is clear that, irrespective of the time step and block, the spatial token scorer assigns higher scores to tokens representing the main outline, followed by those representing specific details, and lastly, background tokens.
In terms of temporal scores, the temporal token scorer further extracts feature information by assigning higher scores to tokens that correspond to specific detail features. For example, in the 1st block at $t = 3$, tokens representing the bird's claws receive high scores. In the 2nd and 7th blocks at $t = 3$, tokens representing the head, wings, and branches are assigned high scores. Similarly, at $t = 4$ in the 7th block, tokens representing the bird's tail are given high scores.

The above visualization results can be explained by the design of the spatial token scorer and the temporal token scorer. The spatial token scorer relies on the similarity with neighboring tokens, assigning higher scores to tokens that differ significantly from their local surroundings. This is why the spatial token scorer extracts more information representing texture and boundary features. The computation of the temporal token scorer is inspired by the working mechanism of the human visual system. Specifically, tokens representing distinct features are captured over time, while background tokens are gradually ignored. As a result, the temporal token scorer is able to extract specific feature information. Overall, by combining the spatial token scorer with the temporal token scorer, IRToP effectively selects informative tokens, reducing computational resources while maintaining performance.


\section{Why cannot direct token pruning be extended to the feature variant spiking transformer?}
\label{sec:why}



Direct token pruning refers to identifying uninformative tokens and discarding them without further processing in the subsequent network layers. In this section, we first discuss why this approach cannot be applied to feature-variant spiking transformers. Then, we use a feature-variant QKFormer as an example to illustrate the issue. Finally, we explain how our method effectively solves this issue.

Early SNN transformers, e.g., Spikformer \cite{zhouspikformer} and SDT-V1 \cite{yao2023spike}, follow ViT-style designs from ANNs, using patch embedding and standard transformer blocks. As the field developed, recent SOTA models like QKFormer \cite{zhou2024qkformer} and SDT-V2/V3 \cite{yaospike,yao2025scaling} incorporate convolution layers with kernels larger than 1 inside transformer blocks. For example, QKFormer applies conv-based Spiking Patch Embedding before each block, and SDT-V3 uses spike-based separable convolutions before every attention layer. Unlike ANNs where features can be flattened for token pruning, these convolutional layers embedded in the transformer blocks require structured and square feature maps for token pruning in SNNs. 
These feature-variant spiking transformers include many operations that reduce the size of feature maps, such as downsampling and convolution. The inherent structural sensitivity of these operations makes direct token pruning incompatible with feature-variant spiking Transformers, which can be understood from two aspects:
\begin{itemize}
    \item On the one hand, convolution operations rely on structured grid-like inputs. This means that if tokens are removed from a transformer block, the remaining tokens may no longer form a valid image layout, making them incompatible with later convolutional layers. 
    \item On the other hand, due to its strong prior assumptions, e.g., spatial local correlation and translation invariance, the convolution operation heavily relies on the spatial structure of feature maps. However, removing tokens disrupts the spatial structure of feature maps. This disruption (1) impairs local information propagation, (2) degrades the effectiveness of trained filters, and (3) compromises the model's representational ability. 
\end{itemize}
As detailed in the above two reasons, it is the existence of convolution in spiking transformers that makes direct token removal infeasible, while our block-level early stopping strategy remains viable. This also indicates that, when pruning advanced spiking transformers like QKFormer and SDT-V2/V3, it is essential to preserve the overall architectural integrity. Notably, existing SNN token pruning methods have only been tested on ViT-like Spikformer and spike-driven transformer V1, and have not yet been applied to recent SOTA spiking transformers. To the best of our knowledge, we are the first to evaluate token pruning on these advanced spiking architectures.


We then use the advanced hierarchical transformer architecture QKFormer in SNNs as an example to illustrate the above issue. In QKFormer, each stage consists of the Spiking Patch Embedding with Deformed Shortcut (SPEDS) module and QKFormer block. The SPEDS module includes structure-sensitive convolution and pooling operations, which reduce the number of tokens by a 2$\times$2 patch size before each stage and transform the number of channels into 2$C$ to generate hierarchical spiking representations. If we directly prune the uninformative tokens identified by IRToP criterion in the first stage, the remaining informative tokens need be reorganized into a new feature map before being input into the second stage. 
This will lead to the following two challenges.
\begin{itemize}
    \item \textit{Difficulty in reshaping the feature map}. The reorganizing process typically requires the remaining informative tokens to be arranged into a square feature map for efficient processing in the next stage (e.g., 196 = 14×14). However, after pruning uninformative tokens in the first stage, the number of remaining tokens often cannot form the required square shape for reshaping the feature map.

    \item \textit{Disruption of spatial structure}. Even if we constrain the remaining informative tokens' count to match the square shape, the spatial structure of the reconstructed feature map is inevitably disrupted. This results in the failure of well-trained parameters in the convolution operations within the SPEDS module of the second stage. This would affect the information flow in subsequent network layers and significantly degrade model performance.
\end{itemize}

TP-Spikformer addresses this challenge by introducing a block-level early stopping strategy for uninformative tokens. Instead of directly removing tokens that would disrupt the spatial structure of the feature map, TP-Spikformer bypasses the processing of uninformative tokens within the transformer blocks, and then reorganizes all tokens spatially before inputting them into the next stage.
This process reduces the memory and computational overhead associated with token pruning by bypassing the computation of uninformative tokens. Moreover, by preserving the integrity of the feature map’s spatial structure, TP-Spikformer avoids the difficulties of reshaping tokens and maintains the well-trained parameters of the filters, ensuring that the model maintains competitive performance even without fine-tuning.

\section{Zero-finetuning accuracy preservation of TP-Spikformer}

The zero-finetuning accuracy preservation of TP-Spikformer is made in a comparative sense. Existing advanced token pruning methods in SNNs often modify the original model architecture when applied to spiking transformers. These modifications may include introducing new tokens (STATA \cite{zhuge2024towards}), adding trainable modules (ACT \cite{kang2024snn}). Since these additions are randomly initialized, they require full retraining, which significantly increases data requirements, training costs, and reduces generalizability. Therefore, though our method does not completely preserve accuracy on QKFormer and SDT-V3, it achieves better accuracy than existing spiking token pruning methods under the same no-fine-tuning setting.

\begin{wrapfigure}[16]{r}{0.55\textwidth}
\centering
\includegraphics[scale=0.53]{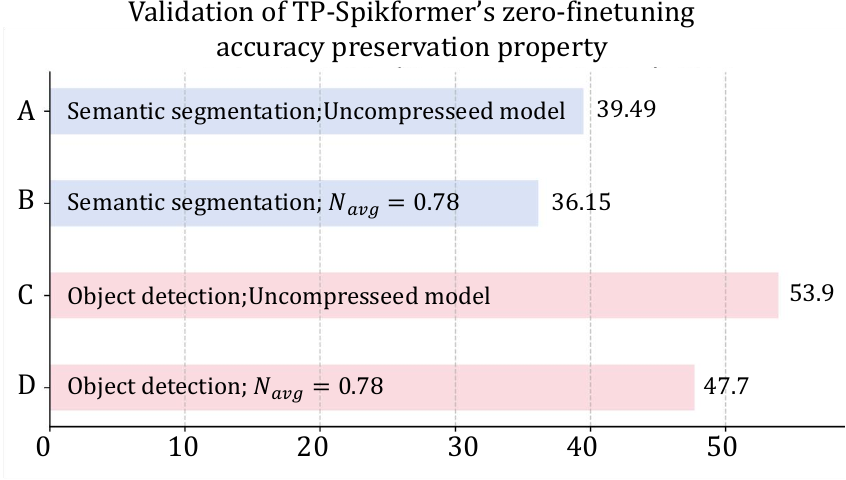}
\vspace{-1mm}
\caption{TP-Spikformer's zero-finetuning accuracy preservation on segmentation and detection.}
\vspace{-8mm}
\label{fig:zerot}
\end{wrapfigure}
In Section \ref{sec:valstu}, we conduct an in-depth analysis of TP-Spikformer's zero-finetuning performance preservation property in image classification tasks. In this section, we demonstrate that TP-Spikformer exhibits this property in other vision tasks as well. We evaluate the performance of the unpruned model using publicly available SDT-V3 detection and segmentation code and obtain results of 39.49\% MIoU for semantic segmentation and 53.9\% mAP@0.5 for object detection, respectively. Then, we directly apply our token pruning method to the obtained weights without any fine-tuning. With a compression ratio of 0.78, TP-Spikformer achieves performance of 36.15\% and 47.7\% in semantic segmentation and object detection, representing reductions of 3.34\% and 6.2\%, respectively. 
These results indicate that TP-Spikformer maintains its zero-finetuning performance preservation property in various downstream vision tasks, highlighting its effectiveness for real-world scenarios with limited resources and no retraining budget.

\section{Effect of TP-Spikformer in accelerating training from scratch}
\begin{table}[htbp]
\centering
\caption{Effect of TP-Spikformer in the training process.}
\label{tab:train}
\begin{tabular}{@{}lccccc@{}}
\toprule
Model & Params (M) & \makecell{Token preservation\\ratio} & \makecell{Total training\\time} & \makecell{GPU Memory\\(batch size=1280)} & Acc. \\
\midrule
SDT-V3 & 5.1 & 1 & 57h14min & 77.49 GB & 73.9\% \\
TP-Spikformer & 5.1 & 0.65 & 50h 49min & 66.90 GB & 73.6\% \\
\bottomrule
\end{tabular}
\end{table}
Although our method is mainly designed to improve deployment efficiency, it can also be easily applied during training. To verify this, we conduct experiments on large-scale ImageNet using SDT-V3-5M with and without our token pruning method, both trained from scratch. All experiments are run on a single H800 GPU, using the same settings as the original SDT-V3 to ensure a fair comparison. As shown in Table \ref{tab:train}, TP-Spikformer achieves 73.6\% accuracy, close to the 73.9\% of the unpruned model, while reducing training time by 7.5 hours. This shows its effectiveness in speeding up training. Moreover, TP-Spikformer uses much less GPU memory due to fewer tokens, which allows larger models or batch sizes to be trained on the same hardware.

\section{Validity of TP-Spikformer on non-visual tasks}
The vision datasets used in the manuscript, like ImageNet, COCO, and ADE20K, are universally recognized as complex datasets in the fields of classification, detection, and segmentation, respectively. These experimental results prove the effectiveness of S$^2$NN in complex image tasks. To further show the efficacy of our method in non-image tasks, we have conducted experiments in NLP tasks. 

We extend our TP-Spikformer to the SpikeLM proposed by \cite{xing2024spikelm} without additional architecture adjustment. Experiments utilize a 12-layer BERT-based encoder transformer and are performed on the GLUE benchmark. The token preservation ratio is set to [1, 1, 1, 0.9, 0.9, 0.9, 0.8, 0.8, 0.8, 0.7, 0.7, 0.7], while all other training configurations follow the original paper. The results are presented in the table below. Clearly, TP-Spikformer achieves a performance of 75.9\%, showing no significant loss. These results confirm the effectiveness of our method on NLP tasks, demonstrating its general applicability beyond the vision domain.
\begin{table}[htbp]
\centering
\caption{Validation of TP-Spikformer on the GLUE benchmark.}
\label{tab:glue_results}
\begin{tabular}{@{}l*{8}{c}c@{}}
\toprule
{Model} & SST-2 & MRPC & RTE & MNLI & QNLI & QQP & CoLA & STS-B  & {Avg.} \\
\midrule
SpikeLM & 87.0 & 85.7 & 69.0 & 77.1 & 85.3 & 83.9 & 38.8 & 84.9 & 76.5 \\
TP-Spikformer & 87.9 & 84.7 & 68.2 & 76.0 & 84.6 & 84.2 & 37.0 & 84.9 & 75.9 \\
\bottomrule
\end{tabular}
\end{table}

\section{Potential of TP-Spikformer with other compression techniques}
Our TP-Spikformer is orthogonal to other lightweight approaches and can be used in conjunction with them.
Here, we investigate the combination of TP-Spikformer with quantization. We select the Q-SDT proposed by \cite{qiu2025quantized} to evaluate this combination. Specifically, we conduct experiments on CIFAR-10 by applying TP-Spikformer to Q-SDT, training the model from scratch. The results are shown in Table \ref{tab:quantization_results}. While there is a performance gap compared to the baseline, the 96.9\% accuracy demonstrates that our token pruning method can work effectively with quantization techniques.
\begin{table}[htbp]
\centering
\caption{Potential of TP-Spikformer with quantization.}
\label{tab:quantization_results}
\begin{tabular}{@{}lccccc@{}}
\toprule
Model & \makecell{Preserving Ratio} & Bits & \makecell{GPU Memory (GB)} & \makecell{Accuracy (\%)} \\
\midrule
Q-SDT \cite{qiu2025quantized} & 1 & 4-1 & 19.27 & 97.8\% \\
Q-SDT+TP-Spikformer & 0.54 & 4-1 & 15.98 & 96.9\% \\
\bottomrule
\end{tabular}
\end{table}

\section{Learning algorithm of TP-Spikformer}
We employ the widely used spatiotemporal backpropagation (STBP) \cite{wu2018spatio,jiang2025mpd} to train the TP-Spikformer, where we need to compute gradients of the loss function $\mathcal{L}$ to synaptic weights. 
Through chain rule decomposition, we can decompose this gradient as,
\begin{equation}
\frac{\partial \mathcal{L}}{\partial \mathbf{w}_{ij}^{\ell}} = \sum_{t=1}^{T} \left( \frac{\partial \mathcal{L}}{\partial \mathbf{s}_{i}^{\ell}[t]} \frac{\partial \mathbf{s}_{i}^{\ell}[t]}{\partial \mathbf{\tilde{u}}_{i}^{\ell}[t]} \frac{\partial \mathbf{\tilde{u}}_{i}^{\ell}[t]}{\partial \mathbf{w}_{ij}^{\ell}} + \frac{\partial \mathcal{L}}{\partial \mathbf{u}_{i}^{\ell}[t+1]} \frac{\partial \mathbf{u}_{i}^{\ell}[t+1]}{\partial \mathbf{\tilde{u}}_{i}^{\ell}[t]} \frac{\partial \mathbf{\tilde{u}}_{i}^{\ell}[t]}{\partial \mathbf{w}_{ij}^{\ell}} \right),   
\end{equation}
where the derivative of the loss function with respect to the spike and membrane potential, i.e., ${\partial \mathcal{L}}/{\partial \mathbf{s}_{i}^{\ell}[t]}$ and ${\partial \mathcal{L}}/{\partial \mathbf{u}_{i}^{\ell}[t+1]}$ are obtained iteratively, the terms of ${\partial \mathbf{\tilde{u}}_{i}^{\ell}[t]}/{\partial \mathbf{w}_{ij}^{\ell}}$, ${\partial \mathbf{u}_{i}^{\ell}[t+1]}/{\partial \mathbf{\tilde{u}}_{i}^{\ell}[t]}$, and ${\partial \mathbf{\tilde{u}}_{i}^{\ell}[t]}/{\partial \mathbf{w}_{ij}^{\ell}}$ can be calculated based on Eq. \ref{eq:mem}.
Unfortunately, a fundamental challenge in this training arises from the non-differentiable nature of spike emission. Mathematically, the gradient of the spike generation function as described in Eq.~\ref{eq:lif}, i.e., $\partial \mathbf{s}_{i}^{\ell}[t] / \partial \mathbf{\tilde{u}}_{i}^{\ell}[t]$, becomes undefined at the firing threshold $\theta$ and vanishes elsewhere. This discontinuity prevents the direct application of standard backpropagation algorithms commonly used in deep learning. To overcome this limitation, we use surrogate gradient functions to approximate the derivative of $\partial \mathbf{s}_{i}^{\ell}[t] / \partial \mathbf{\tilde{u}}_{i}^{\ell}[t]$ \cite{wu2018spatio}, with various functions can be employed like rectangular \cite{wu2019direct}, triangular \cite{deng2022temporal}, and linear \cite{wei2024event}. TP-Spikformer employs the triangular-shaped surrogate gradient formulation, described as,
\begin{equation}
\frac{\partial \mathbf{s}_{i}^{\ell}[t]}{\partial \mathbf{\tilde{u}}_{i}^{\ell}[t]} = \max \left( 0, \beta - |\mathbf{\tilde{u}}_{i}^{\ell}[t] - \theta| \right),
\end{equation}
where \( \beta \) is the factor that defines the range of gradient computation, and \( \theta \) is the threshold as in Eq. \ref{eq:lif}. Consequently, the TP-Spikformer can be trained directly with gradient backpropagation.

\section{Theoretical Energy Consumption}
When analyzing the energy consumption of SNNs, previous studies \cite{yao2023attention, zhouspikformer, zhou2024qkformer,wei2025s,wei2025qp} commonly assume that MAC and AC operations are implemented on 45nm hardware \cite{45nm}, where $E_{MAC} = 4.6{pJ}$ and $E_{AC} = 0.9{pJ}$. To facilitate comparison between different methods, we adopt this approach to theoretically calculate TP-Spikformer's energy consumption, described by the following equation:
\begin{align}
\small
    E_{total}&=E_{MAC}\cdot FLOPs_{Conv}^1 + E_{AC} \times(\sum_{n=2}^N SOPs_{Conv}^n  + \sum_{l=1}^L \times SOPs^l_{Block}+SOPs_{MLP}),
\end{align}
where $SOPs$ refers to the number of synaptic operations, $SOPs_{conv}^n$ and $SOPs_{MLP}^m$ represent the $SOPs$ for the convolutional operations in the embedding module and the MLP in the classification head, respectively, and $SOPs^l_{Block}$ denotes the $SOPs$ for each transformer block. The number of $SOPs$ in TP-Spikformer is computed as:
\begin{align}
    SOPs^\ell&=fr^\ell_{Avg}\times T \times FLOPs^{\ell},
\end{align}
where $fr_{Avg}$ is the average firing rate of the layer across time steps $T$, and $FLOPs^l$ is  the number of floating point operations for the $\ell$-th layer. A spiking transformer typically consists of three components: path embedding, transformer blocks, and the classification head. In TP-Spikformer, the energy consumption of the path embedding and classification head is consistent with the uncompressed counterpart, while the energy consumption of the transformer blocks is significantly reduced.

\section{Instructions for using Large Language Models}
In preparing this manuscript, we utilize a large language model (LLM) solely to aid and polish the writing. The LLM is used for grammar checking, language refinement, and improving clarity of expression. It does not contribute to the formulation of research ideas, methodology, experiments, data analysis, or conclusions. All presented in this paper is entirely the work of the authors.
\end{document}

%% file: math_commands.tex
\usepackage{amsmath,amsfonts,bm}

\def\eqref#1{equation~\ref{#1}}

\def\1{\bm{1}}

\DeclareMathAlphabet{\mathsfit}{\encodingdefault}{\sfdefault}{m}{sl}
\SetMathAlphabet{\mathsfit}{bold}{\encodingdefault}{\sfdefault}{bx}{n}













%% file: iclr2026_conference.bbl
\begin{thebibliography}{79}
\providecommand{\natexlab}[1]{#1}
\providecommand{\url}[1]{\texttt{#1}}
\expandafter\ifx\csname urlstyle\endcsname\relax
  \providecommand{\doi}[1]{doi: #1}\else
  \providecommand{\doi}{doi: \begingroup \urlstyle{rm}\Url}\fi

\bibitem[Akopyan et~al.(2015)Akopyan, Sawada, Cassidy, Alvarez-Icaza, Arthur, Merolla, Imam, Nakamura, Datta, Nam, et~al.]{akopyan2015truenorth}
Filipp Akopyan, Jun Sawada, Andrew Cassidy, Rodrigo Alvarez-Icaza, John Arthur, Paul Merolla, Nabil Imam, Yutaka Nakamura, Pallab Datta, Gi-Joon Nam, et~al.
\newblock Truenorth: Design and tool flow of a 65 mw 1 million neuron programmable neurosynaptic chip.
\newblock \emph{IEEE transactions on computer-aided design of integrated circuits and systems}, 34\penalty0 (10):\penalty0 1537--1557, 2015.

\bibitem[Cai et~al.(2024)Cai, Liu, and Wang]{cai2024hiptrack}
Wenrui Cai, Qingjie Liu, and Yunhong Wang.
\newblock Hiptrack: Visual tracking with historical prompts.
\newblock In \emph{Proceedings of the IEEE/CVF Conference on Computer Vision and Pattern Recognition}, pp.\  19258--19267, 2024.

\bibitem[Cao et~al.(2025)Cao, Zhou, Wei, Belatreche, Liang, Zhang, Zhang, Yang, and Li]{cao2025binary}
Honglin Cao, Zijian Zhou, Wenjie Wei, Ammar Belatreche, Yu~Liang, Dehao Zhang, Malu Zhang, Yang Yang, and Haizhou Li.
\newblock Binary event-driven spiking transformer.
\newblock \emph{arXiv preprint arXiv:2501.05904}, 2025.

\bibitem[Che et~al.(2024)Che, Zhou, Niu, Ma, Fang, Chen, Shen, Yuan, and Tian]{che2024auto}
Kaiwei Che, Zhaokun Zhou, Jun Niu, Zhengyu Ma, Wei Fang, Yanqi Chen, Shuaijie Shen, Li~Yuan, and Yonghong Tian.
\newblock Auto-spikformer: Spikformer architecture search.
\newblock \emph{Frontiers in Neuroscience}, 18:\penalty0 1372257, 2024.

\bibitem[Chen et~al.(2022)Chen, Li, Bai, Qiao, Shen, Li, Gan, Wu, and Ouyang]{chen2022backbone}
Boyu Chen, Peixia Li, Lei Bai, Lei Qiao, Qiuhong Shen, Bo~Li, Weihao Gan, Wei Wu, and Wanli Ouyang.
\newblock Backbone is all your need: A simplified architecture for visual object tracking.
\newblock In \emph{European Conference on Computer Vision}, pp.\  375--392. Springer, 2022.

\bibitem[Chen et~al.(2019)Chen, Wang, Pang, Cao, Xiong, Li, Sun, Feng, Liu, Xu, Zhang, Cheng, Zhu, Cheng, Zhao, Li, Lu, Zhu, Wu, Dai, Wang, Shi, Ouyang, Loy, and Lin]{mmdetection}
Kai Chen, Jiaqi Wang, Jiangmiao Pang, Yuhang Cao, Yu~Xiong, Xiaoxiao Li, Shuyang Sun, Wansen Feng, Ziwei Liu, Jiarui Xu, Zheng Zhang, Dazhi Cheng, Chenchen Zhu, Tianheng Cheng, Qijie Zhao, Buyu Li, Xin Lu, Rui Zhu, Yue Wu, Jifeng Dai, Jingdong Wang, Jianping Shi, Wanli Ouyang, Chen~Change Loy, and Dahua Lin.
\newblock {MMDetection}: Open mmlab detection toolbox and benchmark.
\newblock \emph{arXiv preprint arXiv:1906.07155}, 2019.

\bibitem[Chen et~al.(2023)Chen, Peng, Wang, Lu, and Hu]{chen2023seqtrack}
Xin Chen, Houwen Peng, Dong Wang, Huchuan Lu, and Han Hu.
\newblock Seqtrack: Sequence to sequence learning for visual object tracking.
\newblock In \emph{Proceedings of the IEEE/CVF conference on computer vision and pattern recognition}, pp.\  14572--14581, 2023.

\bibitem[Contributors(2020)]{contributors2020mmsegmentation}
MMSegmentation Contributors.
\newblock Mmsegmentation: Openmmlab semantic segmentation toolbox and benchmark, 2020.

\bibitem[Davies et~al.(2018)Davies, Srinivasa, Lin, Chinya, Cao, Choday, Dimou, Joshi, Imam, Jain, et~al.]{davies2018loihi}
Mike Davies, Narayan Srinivasa, Tsung-Han Lin, Gautham Chinya, Yongqiang Cao, Sri~Harsha Choday, Georgios Dimou, Prasad Joshi, Nabil Imam, Shweta Jain, et~al.
\newblock Loihi: A neuromorphic manycore processor with on-chip learning.
\newblock \emph{Ieee Micro}, 38\penalty0 (1):\penalty0 82--99, 2018.

\bibitem[Deng et~al.(2009)Deng, Dong, Socher, Li, Li, and Fei-Fei]{deng2009imagenet}
Jia Deng, Wei Dong, Richard Socher, Li-Jia Li, Kai Li, and Li~Fei-Fei.
\newblock Imagenet: A large-scale hierarchical image database.
\newblock In \emph{2009 IEEE conference on computer vision and pattern recognition}, pp.\  248--255. Ieee, 2009.

\bibitem[Deng et~al.(2022)Deng, Li, Zhang, and Gu]{deng2022temporal}
Shikuang Deng, Yuhang Li, Shanghang Zhang, and Shi Gu.
\newblock Temporal efficient training of spiking neural network via gradient re-weighting.
\newblock \emph{arXiv preprint arXiv:2202.11946}, 2022.

\bibitem[Devlin et~al.(2019)Devlin, Chang, Lee, and Toutanova]{devlin2019bert}
Jacob Devlin, Ming-Wei Chang, Kenton Lee, and Kristina Toutanova.
\newblock Bert: Pre-training of deep bidirectional transformers for language understanding.
\newblock In \emph{Proceedings of the 2019 conference of the North American chapter of the association for computational linguistics: human language technologies, volume 1 (long and short papers)}, pp.\  4171--4186, 2019.

\bibitem[Dosovitskiy et~al.(2020)Dosovitskiy, Beyer, Kolesnikov, Weissenborn, Zhai, Unterthiner, Dehghani, Minderer, Heigold, Gelly, et~al.]{dosovitskiy2020image}
Alexey Dosovitskiy, Lucas Beyer, Alexander Kolesnikov, Dirk Weissenborn, Xiaohua Zhai, Thomas Unterthiner, Mostafa Dehghani, Matthias Minderer, Georg Heigold, Sylvain Gelly, et~al.
\newblock An image is worth 16x16 words: Transformers for image recognition at scale.
\newblock \emph{arXiv preprint arXiv:2010.11929}, 2020.

\bibitem[Fang et~al.(2021)Fang, Yu, Chen, Huang, Masquelier, and Tian]{fang2021deep}
Wei Fang, Zhaofei Yu, Yanqi Chen, Tiejun Huang, Timoth{\'e}e Masquelier, and Yonghong Tian.
\newblock Deep residual learning in spiking neural networks.
\newblock \emph{Advances in Neural Information Processing Systems}, 34:\penalty0 21056--21069, 2021.

\bibitem[Fecteau \& Munoz(2006)Fecteau and Munoz]{fecteau2006salience}
Jillian~H Fecteau and Douglas~P Munoz.
\newblock Salience, relevance, and firing: a priority map for target selection.
\newblock \emph{Trends in cognitive sciences}, 10\penalty0 (8):\penalty0 382--390, 2006.

\bibitem[Gerstner \& Kistler(2002)Gerstner and Kistler]{gerstner2002spiking}
Wulfram Gerstner and Werner~M Kistler.
\newblock \emph{Spiking neuron models: Single neurons, populations, plasticity}.
\newblock Cambridge university press, 2002.

\bibitem[Glorot \& Bengio(2010)Glorot and Bengio]{glorot2010understanding}
Xavier Glorot and Yoshua Bengio.
\newblock Understanding the difficulty of training deep feedforward neural networks.
\newblock In \emph{Proceedings of the thirteenth international conference on artificial intelligence and statistics}, pp.\  249--256. JMLR Workshop and Conference Proceedings, 2010.

\bibitem[He et~al.(2017)He, Gkioxari, Doll{\'a}r, and Girshick]{he2017mask}
Kaiming He, Georgia Gkioxari, Piotr Doll{\'a}r, and Ross Girshick.
\newblock Mask r-cnn.
\newblock In \emph{Proceedings of the IEEE international conference on computer vision}, pp.\  2961--2969, 2017.

\bibitem[Horowitz(2014)]{45nm}
Mark Horowitz.
\newblock 1.1 computing's energy problem (and what we can do about it).
\newblock In \emph{2014 IEEE International Solid-State Circuits Conference Digest of Technical Papers (ISSCC)}, pp.\  10--14, 2014.
\newblock \doi{10.1109/ISSCC.2014.6757323}.

\bibitem[Itti et~al.(2002)Itti, Koch, and Niebur]{itti2002model}
Laurent Itti, Christof Koch, and Ernst Niebur.
\newblock A model of saliency-based visual attention for rapid scene analysis.
\newblock \emph{IEEE Transactions on pattern analysis and machine intelligence}, 20\penalty0 (11):\penalty0 1254--1259, 2002.

\bibitem[Izhikevich(2003)]{izhikevich2003simple}
Eugene~M Izhikevich.
\newblock Simple model of spiking neurons.
\newblock \emph{IEEE Transactions on neural networks}, 14\penalty0 (6):\penalty0 1569--1572, 2003.

\bibitem[Jiang et~al.(2025)Jiang, Jiang, Yan, and Tang]{jiang2025mpd}
Runhao Jiang, Chengzhi Jiang, Rui Yan, and Huajin Tang.
\newblock Mpd-sgr: Robust spiking neural networks with membrane potential distribution-driven surrogate gradient regularization.
\newblock \emph{arXiv preprint arXiv:2511.12199}, 2025.

\bibitem[Kang et~al.(2023)Kang, Chen, Wang, Peng, and Lu]{kang2023exploring}
Ben Kang, Xin Chen, Dong Wang, Houwen Peng, and Huchuan Lu.
\newblock Exploring lightweight hierarchical vision transformers for efficient visual tracking.
\newblock In \emph{Proceedings of the IEEE/CVF International Conference on Computer Vision}, pp.\  9612--9621, 2023.

\bibitem[Kang et~al.(2024)Kang, Lee, Lee, Kang, Lee, and Baek]{kang2024snn}
Donghwa Kang, Youngmoon Lee, Eun-Kyu Lee, Brent Kang, Jinkyu Lee, and Hyeongboo Baek.
\newblock At-snn: Adaptive tokens for vision transformer on spiking neural network.
\newblock \emph{arXiv preprint arXiv:2408.12293}, 2024.

\bibitem[Kim et~al.(2020)Kim, Park, Na, and Yoon]{kim2020spiking}
Seijoon Kim, Seongsik Park, Byunggook Na, and Sungroh Yoon.
\newblock Spiking-yolo: spiking neural network for energy-efficient object detection.
\newblock In \emph{Proceedings of the AAAI conference on artificial intelligence}, volume~34, pp.\  11270--11277, 2020.

\bibitem[Kirillov et~al.(2019)Kirillov, Girshick, He, and Doll{\'a}r]{kirillov2019panoptic}
Alexander Kirillov, Ross Girshick, Kaiming He, and Piotr Doll{\'a}r.
\newblock Panoptic feature pyramid networks.
\newblock In \emph{Proceedings of the IEEE/CVF conference on computer vision and pattern recognition}, pp.\  6399--6408, 2019.

\bibitem[Koch \& Ullman(1987)Koch and Ullman]{koch1987shifts}
Christof Koch and Shimon Ullman.
\newblock Shifts in selective visual attention: towards the underlying neural circuitry.
\newblock In \emph{Matters of intelligence: Conceptual structures in cognitive neuroscience}, pp.\  115--141. Springer, 1987.

\bibitem[Krizhevsky et~al.(2009)Krizhevsky, Hinton, et~al.]{krizhevsky2009learning}
Alex Krizhevsky, Geoffrey Hinton, et~al.
\newblock Learning multiple layers of features from tiny images.
\newblock 2009.

\bibitem[Law \& Deng(2018)Law and Deng]{law2018cornernet}
Hei Law and Jia Deng.
\newblock Cornernet: Detecting objects as paired keypoints.
\newblock In \emph{Proceedings of the European conference on computer vision (ECCV)}, pp.\  734--750, 2018.

\bibitem[Li et~al.(2024)Li, Deng, Tang, Pan, Tian, Roy, and Maass]{li2024brain}
Guoqi Li, Lei Deng, Huajin Tang, Gang Pan, Yonghong Tian, Kaushik Roy, and Wolfgang Maass.
\newblock Brain-inspired computing: A systematic survey and future trends.
\newblock \emph{Proceedings of the IEEE}, 2024.

\bibitem[Li et~al.(2017)Li, Liu, Ji, Li, and Shi]{li2017cifar10}
Hongmin Li, Hanchao Liu, Xiangyang Ji, Guoqi Li, and Luping Shi.
\newblock Cifar10-dvs: an event-stream dataset for object classification.
\newblock \emph{Frontiers in neuroscience}, 11:\penalty0 244131, 2017.

\bibitem[Li et~al.(2022)Li, He, Dong, Kong, and Zeng]{li2022spike}
Yang Li, Xiang He, Yiting Dong, Qingqun Kong, and Yi~Zeng.
\newblock Spike calibration: Fast and accurate conversion of spiking neural network for object detection and segmentation.
\newblock \emph{arXiv preprint arXiv:2207.02702}, 2022.

\bibitem[Lin et~al.(2021)Lin, Ji, Zhang, Zhang, Wu, and Tian]{lin2021channel}
Mingbao Lin, Rongrong Ji, Yuxin Zhang, Baochang Zhang, Yongjian Wu, and Yonghong Tian.
\newblock Channel pruning via automatic structure search.
\newblock In \emph{Proceedings of the Twenty-Ninth International Conference on International Joint Conferences on Artificial Intelligence}, pp.\  673--679, 2021.

\bibitem[Lin et~al.(2014)Lin, Maire, Belongie, Hays, Perona, Ramanan, Doll{\'a}r, and Zitnick]{lin2014microsoft}
Tsung-Yi Lin, Michael Maire, Serge Belongie, James Hays, Pietro Perona, Deva Ramanan, Piotr Doll{\'a}r, and C~Lawrence Zitnick.
\newblock Microsoft coco: Common objects in context.
\newblock In \emph{Computer Vision--ECCV 2014: 13th European Conference, Zurich, Switzerland, September 6-12, 2014, Proceedings, Part V 13}, pp.\  740--755. Springer, 2014.

\bibitem[Liu et~al.(2024)Liu, Xiao, Li, and Yu]{liu2024sparsespikformer}
Yue Liu, Shanlin Xiao, Bo~Li, and Zhiyi Yu.
\newblock Sparsespikformer: A co-design framework for token and weight pruning in spiking transformer.
\newblock In \emph{ICASSP 2024-2024 IEEE International Conference on Acoustics, Speech and Signal Processing (ICASSP)}, pp.\  6410--6414. IEEE, 2024.

\bibitem[Neftci et~al.(2019)Neftci, Mostafa, and Zenke]{neftci2019surrogate}
Emre~O Neftci, Hesham Mostafa, and Friedemann Zenke.
\newblock Surrogate gradient learning in spiking neural networks: Bringing the power of gradient-based optimization to spiking neural networks.
\newblock \emph{IEEE Signal Processing Magazine}, 36\penalty0 (6):\penalty0 51--63, 2019.

\bibitem[Nothdurft(2000)]{nothdurft2000salience}
Hans-Christoph Nothdurft.
\newblock Salience from feature contrast: additivity across dimensions.
\newblock \emph{Vision research}, 40\penalty0 (10-12):\penalty0 1183--1201, 2000.

\bibitem[Pfeiffer \& Pfeil(2018)Pfeiffer and Pfeil]{pfeiffer2018deep}
Michael Pfeiffer and Thomas Pfeil.
\newblock Deep learning with spiking neurons: opportunities and challenges.
\newblock \emph{Frontiers in neuroscience}, 12, 2018.

\bibitem[Qiu et~al.(2025)Qiu, Zhang, Zhang, Wei, Cao, Guo, Zhu, Shan, Yang, and Li]{qiu2025quantized}
Xuerui Qiu, Malu Zhang, Jieyuan Zhang, Wenjie Wei, Honglin Cao, Junsheng Guo, Rui-Jie Zhu, Yimeng Shan, Yang Yang, and Haizhou Li.
\newblock Quantized spike-driven transformer.
\newblock \emph{arXiv preprint arXiv:2501.13492}, 2025.

\bibitem[Rao et~al.(2021)Rao, Zhao, Liu, Lu, Zhou, and Hsieh]{rao2021dynamicvit}
Yongming Rao, Wenliang Zhao, Benlin Liu, Jiwen Lu, Jie Zhou, and Cho-Jui Hsieh.
\newblock Dynamicvit: Efficient vision transformers with dynamic token sparsification.
\newblock \emph{Advances in neural information processing systems}, 34:\penalty0 13937--13949, 2021.

\bibitem[Rensink(2002)]{rensink2002change}
Ronald~A Rensink.
\newblock Change detection.
\newblock \emph{Annual review of psychology}, 53\penalty0 (1):\penalty0 245--277, 2002.

\bibitem[Rezatofighi et~al.(2019)Rezatofighi, Tsoi, Gwak, Sadeghian, Reid, and Savarese]{rezatofighi2019generalized}
Hamid Rezatofighi, Nathan Tsoi, JunYoung Gwak, Amir Sadeghian, Ian Reid, and Silvio Savarese.
\newblock Generalized intersection over union: A metric and a loss for bounding box regression.
\newblock In \emph{Proceedings of the IEEE/CVF conference on computer vision and pattern recognition}, pp.\  658--666, 2019.

\bibitem[Roy et~al.(2019)Roy, Jaiswal, and Panda]{roy2019towards}
Kaushik Roy, Akhilesh Jaiswal, and Priyadarshini Panda.
\newblock Towards spike-based machine intelligence with neuromorphic computing.
\newblock \emph{Nature}, 575\penalty0 (7784):\penalty0 607--617, 2019.

\bibitem[Shan et~al.(2025)Shan, Ren, Wu, Wei, Zhu, Wang, Zhang, Xiao, Zhang, Shi, et~al.]{shan2025sdtrack}
Yimeng Shan, Zhenbang Ren, Haodi Wu, Wenjie Wei, Rui-Jie Zhu, Shuai Wang, Dehao Zhang, Yichen Xiao, Jieyuan Zhang, Kexin Shi, et~al.
\newblock Sdtrack: A baseline for event-based tracking via spiking neural networks.
\newblock \emph{arXiv preprint arXiv:2503.08703}, 2025.

\bibitem[Shi et~al.(2024)Shi, Hao, and Yu]{shi2024spikingresformer}
Xinyu Shi, Zecheng Hao, and Zhaofei Yu.
\newblock Spikingresformer: bridging resnet and vision transformer in spiking neural networks.
\newblock In \emph{Proceedings of the IEEE/CVF Conference on Computer Vision and Pattern Recognition}, pp.\  5610--5619, 2024.

\bibitem[Su et~al.(2023)Su, Chou, Hu, Li, Mei, Zhang, and Li]{su2023deep}
Qiaoyi Su, Yuhong Chou, Yifan Hu, Jianing Li, Shijie Mei, Ziyang Zhang, and Guoqi Li.
\newblock Deep directly-trained spiking neural networks for object detection.
\newblock In \emph{Proceedings of the IEEE/CVF International Conference on Computer Vision}, pp.\  6555--6565, 2023.

\bibitem[Wang et~al.(2025)Wang, Zhang, Zhang, Belatreche, Xiao, Liang, Shan, Sun, Zhang, and Yang]{wang2025spiking}
Shuai Wang, Malu Zhang, Dehao Zhang, Ammar Belatreche, Yichen Xiao, Yu~Liang, Yimeng Shan, Qian Sun, Enqi Zhang, and Yang Yang.
\newblock Spiking vision transformer with saccadic attention.
\newblock \emph{arXiv preprint arXiv:2502.12677}, 2025.

\bibitem[Wang et~al.(2023)Wang, Li, Zhu, Zhang, Chen, Li, Wang, Tian, and Wu]{wang2023visevent}
Xiao Wang, Jianing Li, Lin Zhu, Zhipeng Zhang, Zhe Chen, Xin Li, Yaowei Wang, Yonghong Tian, and Feng Wu.
\newblock Visevent: Reliable object tracking via collaboration of frame and event flows.
\newblock \emph{IEEE Transactions on Cybernetics}, 2023.

\bibitem[Wang et~al.(2024{\natexlab{a}})Wang, Huang, Wang, Tang, Jiang, Tian, Tang, and Luo]{wang2024long}
Xiao Wang, Ju~Huang, Shiao Wang, Chuanming Tang, Bo~Jiang, Yonghong Tian, Jin Tang, and Bin Luo.
\newblock Long-term frame-event visual tracking: Benchmark dataset and baseline.
\newblock \emph{arXiv preprint arXiv:2403.05839}, 2024{\natexlab{a}}.

\bibitem[Wang et~al.(2024{\natexlab{b}})Wang, Zhao, Cui, Liu, and Xu]{wang2024autost}
Ziqing Wang, Qidong Zhao, Jinku Cui, Xu~Liu, and Dongkuan Xu.
\newblock Autost: training-free neural architecture search for spiking transformers.
\newblock In \emph{ICASSP 2024-2024 IEEE International Conference on Acoustics, Speech and Signal Processing (ICASSP)}, pp.\  3455--3459. IEEE, 2024{\natexlab{b}}.

\bibitem[Wei et~al.(2024{\natexlab{a}})Wei, Liang, Belatreche, Xiao, Cao, Ren, Wang, Zhang, and Yang]{wei2024q}
Wenjie Wei, Yu~Liang, Ammar Belatreche, Yichen Xiao, Honglin Cao, Zhenbang Ren, Guoqing Wang, Malu Zhang, and Yang Yang.
\newblock Q-snns: Quantized spiking neural networks.
\newblock \emph{arXiv preprint arXiv:2406.13672}, 2024{\natexlab{a}}.

\bibitem[Wei et~al.(2024{\natexlab{b}})Wei, Zhang, Zhang, Belatreche, Wu, Xu, Qiu, Chen, Yang, and Li]{wei2024event}
Wenjie Wei, Malu Zhang, Jilin Zhang, Ammar Belatreche, Jibin Wu, Zijing Xu, Xuerui Qiu, Hong Chen, Yang Yang, and Haizhou Li.
\newblock Event-driven learning for spiking neural networks.
\newblock \emph{arXiv preprint arXiv:2403.00270}, 2024{\natexlab{b}}.

\bibitem[Wei et~al.(2025{\natexlab{a}})Wei, Zhang, Zhang, Belatreche, Wang, Shan, Liu, Cao, Wang, Yang, et~al.]{wei2025s}
Wenjie Wei, Malu Zhang, Jieyuan Zhang, Ammar Belatreche, Shuai Wang, Yimeng Shan, Hanwen Liu, Honglin Cao, Guoqing Wang, Yang Yang, et~al.
\newblock S $^2$ nn: Sub-bit spiking neural networks.
\newblock \emph{arXiv preprint arXiv:2509.24266}, 2025{\natexlab{a}}.

\bibitem[Wei et~al.(2025{\natexlab{b}})Wei, Zhang, Zhou, Belatreche, Shan, Liang, Cao, Zhang, and Yang]{wei2025qp}
Wenjie Wei, Malu Zhang, Zijian Zhou, Ammar Belatreche, Yimeng Shan, Yu~Liang, Honglin Cao, Jieyuan Zhang, and Yang Yang.
\newblock Qp-snn: Quantized and pruned spiking neural networks.
\newblock \emph{arXiv preprint arXiv:2502.05905}, 2025{\natexlab{b}}.

\bibitem[Wei et~al.(2023)Wei, Bai, Zheng, Shi, and Gong]{wei2023autoregressive}
Xing Wei, Yifan Bai, Yongchao Zheng, Dahu Shi, and Yihong Gong.
\newblock Autoregressive visual tracking.
\newblock In \emph{Proceedings of the IEEE/CVF Conference on Computer Vision and Pattern Recognition}, pp.\  9697--9706, 2023.

\bibitem[Wu et~al.(2018)Wu, Deng, Li, Zhu, and Shi]{wu2018spatio}
Yujie Wu, Lei Deng, Guoqi Li, Jun Zhu, and Luping Shi.
\newblock Spatio-temporal backpropagation for training high-performance spiking neural networks.
\newblock \emph{Frontiers in neuroscience}, 12:\penalty0 331, 2018.

\bibitem[Wu et~al.(2019)Wu, Deng, Li, Zhu, Xie, and Shi]{wu2019direct}
Yujie Wu, Lei Deng, Guoqi Li, Jun Zhu, Yuan Xie, and Luping Shi.
\newblock Direct training for spiking neural networks: Faster, larger, better.
\newblock In \emph{Proceedings of the AAAI conference on artificial intelligence}, volume~33, pp.\  1311--1318, 2019.

\bibitem[Xing et~al.(2024)Xing, Zhang, Ni, Xiao, Ju, Fan, Wang, Zhang, and Li]{xing2024spikelm}
Xingrun Xing, Zheng Zhang, Ziyi Ni, Shitao Xiao, Yiming Ju, Siqi Fan, Yequan Wang, Jiajun Zhang, and Guoqi Li.
\newblock Spikelm: Towards general spike-driven language modeling via elastic bi-spiking mechanisms.
\newblock \emph{arXiv preprint arXiv:2406.03287}, 2024.

\bibitem[Yan et~al.(2021)Yan, Peng, Fu, Wang, and Lu]{yan2021learning}
Bin Yan, Houwen Peng, Jianlong Fu, Dong Wang, and Huchuan Lu.
\newblock Learning spatio-temporal transformer for visual tracking.
\newblock In \emph{Proceedings of the IEEE/CVF international conference on computer vision}, pp.\  10448--10457, 2021.

\bibitem[Yao et~al.()Yao, Hu, Hu, Xu, Zhou, Tian, Bo, and Li]{yaospike}
Man Yao, JiaKui Hu, Tianxiang Hu, Yifan Xu, Zhaokun Zhou, Yonghong Tian, XU~Bo, and Guoqi Li.
\newblock Spike-driven transformer v2: Meta spiking neural network architecture inspiring the design of next-generation neuromorphic chips.
\newblock In \emph{The Twelfth International Conference on Learning Representations}.

\bibitem[Yao et~al.(2023{\natexlab{a}})Yao, Hu, Zhou, Yuan, Tian, Xu, and Li]{yao2023spike}
Man Yao, Jiakui Hu, Zhaokun Zhou, Li~Yuan, Yonghong Tian, Bo~Xu, and Guoqi Li.
\newblock Spike-driven transformer.
\newblock \emph{Advances in neural information processing systems}, 36:\penalty0 64043--64058, 2023{\natexlab{a}}.

\bibitem[Yao et~al.(2023{\natexlab{b}})Yao, Zhao, Zhang, Hu, Deng, Tian, Xu, and Li]{yao2023attention}
Man Yao, Guangshe Zhao, Hengyu Zhang, Yifan Hu, Lei Deng, Yonghong Tian, Bo~Xu, and Guoqi Li.
\newblock Attention spiking neural networks.
\newblock \emph{IEEE transactions on pattern analysis and machine intelligence}, 2023{\natexlab{b}}.

\bibitem[Yao et~al.(2025)Yao, Qiu, Hu, Hu, Chou, Tian, Liao, Leng, Xu, and Li]{yao2025scaling}
Man Yao, Xuerui Qiu, Tianxiang Hu, Jiakui Hu, Yuhong Chou, Keyu Tian, Jianxing Liao, Luziwei Leng, Bo~Xu, and Guoqi Li.
\newblock Scaling spike-driven transformer with efficient spike firing approximation training.
\newblock \emph{IEEE Transactions on Pattern Analysis and Machine Intelligence}, 2025.

\bibitem[Ye et~al.(2022)Ye, Chang, Ma, Shan, and Chen]{ye2022joint}
Botao Ye, Hong Chang, Bingpeng Ma, Shiguang Shan, and Xilin Chen.
\newblock Joint feature learning and relation modeling for tracking: A one-stream framework.
\newblock In \emph{European Conference on Computer Vision}, pp.\  341--357. Springer, 2022.

\bibitem[Yin et~al.(2022)Yin, Vahdat, Alvarez, Mallya, Kautz, and Molchanov]{yin2022vit}
Hongxu Yin, Arash Vahdat, Jose~M Alvarez, Arun Mallya, Jan Kautz, and Pavlo Molchanov.
\newblock A-vit: Adaptive tokens for efficient vision transformer.
\newblock In \emph{Proceedings of the IEEE/CVF conference on computer vision and pattern recognition}, pp.\  10809--10818, 2022.

\bibitem[Zhan et~al.(2025)Zhan, Cao, Xie, Tang, Zhang, Yang, and Liu]{zhan2025sfedca}
Qiugang Zhan, Jinbo Cao, Xiurui Xie, Huajin Tang, Malu Zhang, Shantian Yang, and Guisong Liu.
\newblock Sfedca: Credit assignment-based active client selection strategy for spiking federated learning.
\newblock \emph{IEEE Transactions on Neural Networks and Learning Systems}, 2025.

\bibitem[Zhang et~al.(2024)Zhang, Zhou, Yu, Huang, Fan, Yuan, Ma, Zhou, Tian, et~al.]{zhang2024qkformer}
Han Zhang, Zhaokun Zhou, Liutao Yu, Liwei Huang, Xiaopeng Fan, Li~Yuan, Zhengyu Ma, Huihui Zhou, Yonghong Tian, et~al.
\newblock Qkformer: Hierarchical spiking transformer using qk attention.
\newblock \emph{Advances in Neural Information Processing Systems}, 37:\penalty0 13074--13098, 2024.

\bibitem[Zhang et~al.(2023)Zhang, Li, He, Fan, Wang, and Zhang]{zhang2023direct}
Hong Zhang, Yang Li, Bin He, Xiongfei Fan, Yue Wang, and Yu~Zhang.
\newblock Direct training high-performance spiking neural networks for object recognition and detection.
\newblock \emph{Frontiers in Neuroscience}, 17:\penalty0 1229951, 2023.

\bibitem[Zhang et~al.(2021{\natexlab{a}})Zhang, Yang, Fu, Wei, Yin, and Dong]{zhang2021object}
Jiqing Zhang, Xin Yang, Yingkai Fu, Xiaopeng Wei, Baocai Yin, and Bo~Dong.
\newblock Object tracking by jointly exploiting frame and event domain.
\newblock In \emph{Proceedings of the IEEE/CVF International Conference on Computer Vision}, pp.\  13043--13052, 2021{\natexlab{a}}.

\bibitem[Zhang et~al.(2022)Zhang, Dong, Zhang, Ding, Heide, Yin, and Yang]{zhang2022spiking}
Jiqing Zhang, Bo~Dong, Haiwei Zhang, Jianchuan Ding, Felix Heide, Baocai Yin, and Xin Yang.
\newblock Spiking transformers for event-based single object tracking.
\newblock In \emph{Proceedings of the IEEE/CVF conference on Computer Vision and Pattern Recognition}, pp.\  8801--8810, 2022.

\bibitem[Zhang et~al.(2021{\natexlab{b}})Zhang, Wang, Wu, Belatreche, Amornpaisannon, Zhang, Miriyala, Qu, Chua, Carlson, et~al.]{zhang2021rectified}
Malu Zhang, Jiadong Wang, Jibin Wu, Ammar Belatreche, Burin Amornpaisannon, Zhixuan Zhang, Venkata Pavan~Kumar Miriyala, Hong Qu, Yansong Chua, Trevor~E Carlson, et~al.
\newblock Rectified linear postsynaptic potential function for backpropagation in deep spiking neural networks.
\newblock \emph{IEEE transactions on neural networks and learning systems}, 33\penalty0 (5):\penalty0 1947--1958, 2021{\natexlab{b}}.

\bibitem[Zhang et~al.(2025)Zhang, Wei, Zhou, Liu, Zhang, Belatreche, and Yang]{zhang2025spike}
Malu Zhang, Wenjie Wei, Zijian Zhou, Wanlong Liu, Jie Zhang, Ammar Belatreche, and Yang Yang.
\newblock Spike-driven lightweight large language model with evolutionary computation.
\newblock \emph{IEEE Transactions on Evolutionary Computation}, 2025.

\bibitem[Zheng et~al.(2024)Zheng, Zhong, Liang, Mo, Zhang, and Li]{zheng2024odtrack}
Yaozong Zheng, Bineng Zhong, Qihua Liang, Zhiyi Mo, Shengping Zhang, and Xianxian Li.
\newblock Odtrack: Online dense temporal token learning for visual tracking.
\newblock In \emph{Proceedings of the AAAI Conference on Artificial Intelligence}, volume~38, pp.\  7588--7596, 2024.

\bibitem[Zhou et~al.(2019)Zhou, Zhao, Puig, Xiao, Fidler, Barriuso, and Torralba]{zhou2019semantic}
Bolei Zhou, Hang Zhao, Xavier Puig, Tete Xiao, Sanja Fidler, Adela Barriuso, and Antonio Torralba.
\newblock Semantic understanding of scenes through the ade20k dataset.
\newblock \emph{International Journal of Computer Vision}, 127\penalty0 (3):\penalty0 302--321, 2019.

\bibitem[Zhou et~al.(2024{\natexlab{a}})Zhou, Zhang, Zhou, Yu, Huang, Fan, Yuan, Ma, Zhou, and Tian]{zhou2024qkformer}
Chenlin Zhou, Han Zhang, Zhaokun Zhou, Liutao Yu, Liwei Huang, Xiaopeng Fan, Li~Yuan, Zhengyu Ma, Huihui Zhou, and Yonghong Tian.
\newblock {QKF}ormer: Hierarchical spiking transformer using q-k attention.
\newblock In \emph{The Thirty-eighth Annual Conference on Neural Information Processing Systems}, 2024{\natexlab{a}}.
\newblock URL \url{https://openreview.net/forum?id=AVd7DpiooC}.

\bibitem[Zhou et~al.()Zhou, Zhu, He, Wang, Shuicheng, Tian, and Yuan]{zhouspikformer}
Zhaokun Zhou, Yuesheng Zhu, Chao He, Yaowei Wang, YAN Shuicheng, Yonghong Tian, and Li~Yuan.
\newblock Spikformer: When spiking neural network meets transformer.
\newblock In \emph{The Eleventh International Conference on Learning Representations}.

\bibitem[Zhou et~al.(2024{\natexlab{b}})Zhou, Che, Fang, Tian, Zhu, Yan, Tian, and Yuan]{zhou2024spikformer}
Zhaokun Zhou, Kaiwei Che, Wei Fang, Keyu Tian, Yuesheng Zhu, Shuicheng Yan, Yonghong Tian, and Li~Yuan.
\newblock Spikformer v2: Join the high accuracy club on imagenet with an snn ticket.
\newblock \emph{arXiv preprint arXiv:2401.02020}, 2024{\natexlab{b}}.

\bibitem[Zhou et~al.(2024{\natexlab{c}})Zhou, Che, Niu, Yao, Li, Yuan, Luo, and Zhu]{zhou2024spatial}
Zhaokun Zhou, Kaiwei Che, Jun Niu, Man Yao, Guoqi Li, Li~Yuan, Guibo Luo, and Yuesheng Zhu.
\newblock Spatial-temporal spiking feature pruning in spiking transformer.
\newblock \emph{IEEE Transactions on Cognitive and Developmental Systems}, 2024{\natexlab{c}}.

\bibitem[Zhuge et~al.(2024)Zhuge, Wang, Yao, and Cheng]{zhuge2024towards}
Zhengyang Zhuge, Peisong Wang, Xingting Yao, and Jian Cheng.
\newblock Towards efficient spiking transformer: a token sparsification framework for training and inference acceleration.
\newblock In \emph{Forty-first International Conference on Machine Learning}, 2024.

\end{thebibliography}
